\documentclass[journal]{IEEEtran}

\usepackage{pifont,amssymb,url}
\usepackage{multicol,mathrsfs}
\usepackage[usenames,dvipsnames,svgnames,table]{xcolor}
\usepackage{bm,bbm,mathrsfs,amssymb,pifont,amssymb,pifont,amsfonts,amsmath,stmaryrd,color,amssymb,graphics,graphicx,epstopdf}
\usepackage{amscd,graphicx,array,dsfont,texdraw,tikz,enumerate,multicol,multirow,verbatim}
\usetikzlibrary{arrows,shapes,shadows,positioning,automata,patterns}
\usetikzlibrary{matrix,arrows,calc,trees,decorations.pathmorphing,decorations.markings, decorations.pathreplacing}
\usepackage[utf8x]{inputenc}
\usepackage[T1]{fontenc}
\usepackage{booktabs}

\usetikzlibrary{arrows.meta}
\usetikzlibrary{decorations.pathreplacing}
\newcommand{\tabincell}[2]{\begin{tabular}{@{}#1@{}}#2\end{tabular}}
\renewcommand{\t}{T}

\newcommand{\V}{\mathcal{V}}
\newcommand{\E}{\mathcal{E}}
\newcommand{\G}{\mathcal{G}}
\newcommand{\N}{\mathcal{N}}

\newcommand{\R}{\mathbb{R}}

\newcommand{\col}{\mbox{col}}
\newcommand{\nnum}{\nonumber}

\newcommand{\bremark}{\smallskip\begin{remark}\begin{rm}}
\newcommand{\eremark}{\end{rm}\hfill$\boxempty$\end{remark}\smallskip}
\newcommand{\btheorem}{\smallskip\begin{theorem} \begin{it}}
\newcommand{\etheorem}{\end{it}\end{theorem}\smallskip}
\newcommand{\blemma}{\smallskip\begin{lemma} \begin{it} }
\newcommand{\elemma}{\end{it}\end{lemma}\smallskip}
\newcommand{\bcorollary}{\smallskip\begin{corollary} \begin{it} }
\newcommand{\ecorollary}{\end{it}\end{corollary}\smallskip}
\newcommand{\bdefinition}{\smallskip\begin{definition}\begin{rm}}
\newcommand{\edefinition}{\end{rm}\hfill$\boxempty$\end{definition}\smallskip}
\newcommand{\bproposition}{\smallskip\begin{proposition}\begin{it}}
\newcommand{\eproposition}{\end{it}\end{proposition}\smallskip }
\newcommand{\bexample}{\smallskip\begin{example}\begin{rm}}
\newcommand{\eexample}{\end{rm}\hfill$\boxempty$\end{example}\smallskip}
\newcommand{\bproblem}{\smallskip\begin{problem}\begin{rm}}
\newcommand{\eproblem}{\end{rm}\hfill$\boxempty$\end{problem}\smallskip}

\newcommand{\bassume}{\smallskip\begin{assumption}\begin{rm}}
\newcommand{\eassume}{ \end{rm}\hfill$\boxempty$\end{assumption}\smallskip}
\newcommand{\bfact}{\smallskip\begin{fact}\begin{it}}
\newcommand{\efact}{ \end{it} \end{fact}\smallskip}
\newcommand{\bcondition}{\begin{condition}\begin{rm}}
\newcommand{\econdition}{ \end{rm}\hfill$\boxempty$\end{condition}\smallskip}

\newtheorem{theorem}{Theorem}[section]
\newtheorem{lemma}{Lemma}[section]
\newtheorem{corollary}{Corollary}[section]
\newtheorem{definition}{Definition}[section]
\newtheorem{proposition}{Proposition}[section]
\newtheorem{problem}{Problem}[section]

\newtheorem{myremark}{Remark}[section]
\newenvironment{remark}{\begin{myremark}\normalfont}{\end{myremark}}
\newtheorem{myexample}{Example}[section]
\newenvironment{example}{\begin{myexample}\normalfont}{\end{myexample}}
\newtheorem{assumption}{Assumption}
\newtheorem{fact}{Fact}[section]
\newtheorem{condition}{\it Condition}[section]

\newcommand{\bbm}[1]{\left[\begin{matrix} #1 \end{matrix}\right]}

\newcommand{\bproof}{ \textit{Proof:} \begin{rm} }
\newcommand{\eproof}{ \end{rm} \hfill Q.E.D.}  

\begin{document}
\title{Distributed formation control for manipulator end-effectors}
\author{Haiwen Wu, Bayu Jayawardhana, Hector Garcia de Marina and Dabo Xu %
\thanks{Haiwen Wu and Bayu Jayawardhana are with Engineering and Technology Institute Groningen, Faculty of Science and Engineering, University of Groningen, Groningen 9747 AG, The Netherlands (e-mails: haiwen.wu@rug.nl; b.jayawardhana@rug.nl).
H.G. de Marina is with Department of Computer Architecture and Automatic Control, Faculty of Physics, Universidad Complutense de Madrid, 28040 Madrid, Spain (email: hgarciad@ucm.es). The work of H.G. de Marina is supported by the grant Atraccion de Talento 2019-T2/TIC-13503 from the Goverment of Madrid.
Dabo Xu is with the School of Automation, Nanjing University of Science and Technology, Nanjing 210094, China (e-mail: dxu@njust.edu.cn). The work of D. Xu is supported by National Natural Science Foundation of China under Grant No.~62073168.
}
}


\maketitle

\begin{abstract}
We present three classes of distributed formation controllers for achieving and maintaining the 2D/3D formation shape of manipulator end-effectors to cope with different scenarios due to availability of modeling parameters. 
We firstly present a distributed formation controller for manipulators whose system parameters are perfectly known. 
The formation control objective is achieved by assigning virtual springs between end-effectors and by adding damping terms at joints, which provides a clear physical interpretation of the proposed solution.
Subsequently, we extend it 
to the case where manipulator kinematic and system parameters are not exactly known. An extra integrator and an adaptive estimator are introduced for gravitational compensation and stabilization, respectively.
Simulation results with planar manipulators and with seven degree-of-freedom humanoid manipulator arms are presented to illustrate the effectiveness of the proposed approach.

\end{abstract}



\IEEEpeerreviewmaketitle


\section{Introduction}


This paper investigates the problem of distributed formation control of manipulator end-effectors. Specifically, we consider a group of manipulators whose end-effectors must reach and maintain a prescribed shape in order to fulfill a given group task, such as, collaborative pick-and-place or transportation of large payload, among others. We present distributed control algorithms to solve the problem where the popular gradient-descent formation control for single integrator agents is combined with a passivity-based manipulator controller in the task-space. 

The use of coordinated manipulators or mobile manipulators\footnote{Mobile manipulators refer to mobile robots where manipulator arms are mounted on the mobile platform.} have been developed and deployed in smart manufacturing and logistics systems for the past decades. In these application areas, maintaining a robust formation of robots is important, in particular, when they are used to transport large payloads where slight deformation on the formation can be hazardous. In this context, distributed formation controller can be deployed to the group of robots where each robot uses local on-board sensor systems to maintain formation shape constraints that are defined between the robot and its neighbors \cite{Marina-book,Oh2015survey}. When the agent is considered as a kinematic point (or point mass) whose dynamics is given by single-integrators and double-integrators, fundamental gradient-based control laws have been proposed and studied, for instance, in \cite{chan2021,Oh2014distance,Marina2015controlling,Marina2018taming,zhao2015bearing} where different sources of local information (relative position, distance, bearing or vision) are used. The extension of the formation control to other classes of nonlinear systems includes the formation of non-holonomic wheeled robots \cite{Jafarian2016disturbance,vos2016formation}, spacecraft formation flying \cite{Scharf2003survey,Scharf2004survey}, and dynamic positioning of multiple offshore vessels \cite{Xu2017formation}. In all of these works, the control input acts directly on the state variables that define the formation. It remains an open problem on the design of formation control for systems where the control input does not act directly on the formation error variables, such as, the formation control of underactuated systems or end-effector manipulator systems as studied in this paper.

For the latter case, where we are dealing with the formation control problem for end-effectors, 
the desired formation shape is defined by the end-effectors' position while the control inputs or the actuators act at the joints' level, which makes this problem more challenging. One can consider two level of controllers where distributed formation control law is designed for the formation keeping of end-effectors as kinematic points, and subsequently, the computed velocity at each end-effector for maintaining the formation is back-propagated to the control inputs at the joints' level via inverse kinematics. This multi-level control scheme is, in practice, non-trivial since there is no time-scale separation in the use of collaborative manipulators for high-speed robotization in industry, and the computation of inverse kinematics is computationally demanding.

In our first main result, we start with presenting a gradient-based distributed control design for manipulators whose system parameters are perfectly known based only on local information, and where the desired formation shape can be made exponentially stable for single-integrator agents. The proposed controller is composed of the use of 
virtual springs between the end-effectors and 
damping terms at the joints. Such physics-based control design approach allows us to obtain physical interpretation of the proposed approach. The virtual springs embody the generalization of gradient-based distributed formation control law from the single-integrator agents to the robotic manipulator ones where the distributed control forces for the end-effectors in order to reach and to maintain the formation are distributed to the control forces at each joint.


In our second main result, we extend the gradient-based control law to the case where the manipulator kinematic and dynamic parameters are not exactly known.
Based on the internal model principle, an additional integrator is introduced for gravitational compensation in dealing with uncertainties in the forces coming from the potential energy. To handle uncertainties relating to the kinematics, we  present firstly a controller design using nominal (approximate) Jacobian. A sufficient condition is given to show that the desired shape can be made exponentially stable if the mismatch between the nominal Jacobian and the actual Jacobian is bounded by a known constant. Based on that, we propose an adaptive Jacobian controller which removes the bounded mismatch condition.
Our proposed distributed formation control law uses local information that comes from the on-board sensor systems defined on local coordinate frame. In other words, the relative information of an end-effector's position with respect to its neighbors and the joints' position/velocity of the robot is independent of its neighbors' frames.


The rest of this paper is organized as follows.
In Section~\ref{sec-form}, we present system models, some preliminaries on formation graph and problem formulation.
Our first gradient-based controller for manipulators whose system parameters are exactly known with stability analysis for closed-loop systems is discussed in Section~\ref{sec-grad}.
In Section~\ref{sec-robust}, the extension of the aforementioned gradient-based controller for manipulators with kinematic and dynamic uncertainties is presented. It is followed by further discussions in
Section~\ref{sec-disc}. 
For illustrating the efficacy of our proposed distributed formation control approaches, we show numerical examples in Section~\ref{sec-sim}. 
Concluding remarks are given in Section~\ref{sec-con}. 

\section{Preliminaries and Problem Formulation}\label{sec-form}

\textit{Notation.}
$\|\cdot\|$ is the Euclidean norm in $\R^{n}$ or the induced matrix 2-norm in $\R^{n\times m}$.
For a matrix $A\in\R^{m\times n}$, $A^{\t}$ denotes its transpose, $\lambda_{\min}\{A\}$ and $\lambda_{\max}\{A\}$ denote the minimum and maximum eigenvalue of matrix $A$, respectively.
For column vectors $x_{1},\dots,x_{n}$, we write $\col(x_{1},\dots,x_{n}) := [x_{1}^{\t},\dots,x_{n}^{\t}]^{\t}$ as the stacked column vector.
We will denote by $\otimes$ the Kronecker product, and
we will use a shorthand notation $\overline B := B \otimes I_m$ for any $B\in\R^{n\times m}$ and identity matrix $I_m$ of dimension $m$.

\subsection{Manipulator dynamics and kinematics}
Consider a group of $n$-DOF fully-actuated rigid robotic manipulator modeled by \cite{Slotine-book,Murray1994,Spong-book}
\begin{equation}\label{sys-dyna}
\begin{aligned}
H_{i}(q_{i},w_{i})\ddot{q}_{i} + C_{i}(q_{i},\dot{q}_{i},w_{i})\dot{q}_{i} &+ G_{i}(q_{i},w_{i}) = u_{i}
\end{aligned}
\end{equation}
where $i\in\{1,\ldots,N\}$, where $q_{i}(t),\dot{q}_{i}(t),\ddot{q}_{i}(t)\in\R^{n}$ are the generalized joint position, velocity, and acceleration, respectively, $u_{i}(t)\in\R^{n}$ is the generalized joint control forces, $w_{i}\in\mathbb{W}_{i}\subset\R^{n_{w}}$ is the constant system parameter vector for known bounded compact set $\mathbb{W}_{i}$, $H_{i}(q_{i},w_{i})\in\R^{n \times n}$ is the inertia matrix, $C_{i}(q_{i},\dot{q}_{i},w_{i})\in\R^{n \times n}$ is the Coriolis and centrifugal force matrix-valued function, and $G_{i}(q_{i},w_{i})\in\R^{n}$ is the gravitational torque.

Let $x_{i}(t)\in\R^{m}$ be the $i$th manipulator end-effector position in the task-space (e.g., Cartesian space with $m\in\{2,3\}$) with respect to the world frame $\Sigma_{g}$ and $m\leq n$.  The end-effector position can be mapped to its generalized joint position via a nonlinear forward kinematics mapping \cite{Murray1994,Spong-book}
\begin{equation}\label{sys-kine}
x_{i} = h_{i}(q_{i},w_{i}) + x_{i0}
\end{equation}
where $h_{i}:\R^{n}\times\R^{n_{w}}\to\R^{m}$ is the mapping from joint-space to task-space, and $x_{i0}\in\R^{m}$ is the position of manipulator base with respect to the world frame $\Sigma_{g}$.

Differentiating \eqref{sys-kine} with respect to time gives the relation between the task-space velocity and joint velocity \cite[pp.~196]{Murray1994}, \cite[pp.~122]{Spong-book}
\begin{equation}\label{sys-J}
\dot{x}_{i} = J_{g,i}(q_{i},w_{i})\dot{q}_{i},~~ J_{g,i}(q_{i},w_{i}) := \frac{\partial h_{i}(q_{i},w_{i})}{\partial q_{i}}
\end{equation}
where $J_{g,i}(q_{i},w_{i})\in\R^{m\times n}$ is called the Jacobian matrix of the forward kinematics.

The present study focuses on manipulators with fixed bases.
Suppose all the manipulators are suitably prepositioned such that their \emph{working spaces} are disjoint.
Regarding the kinematic singularities, we denote $\mathbf{Q}_{i} := \{q_{i}\in\R^{n}: \textnormal{dim}\big(\textnormal{null}(J_{g,i}(q_{i},w_{i}))\big) = 0 \}$ be the set away from kinematic singularities. Then for each manipulator (with system parameter $w_{i}$ and base position $x_{i0}$), we define
\begin{equation*}
\mathcal{W}_{i} := \{ x_{i}\in\R^{m} : x_{i}= h_{i}(q_{i},w_{i}) + x_{i0}, \, q_{i}\in\mathbf{Q}_{i} \}
\end{equation*}
as a subset of its reachable \emph{working spaces}. The entire reachable \emph{working space} for the networked manipulators is given by  $\mathcal{W} := \mathcal{W}_{1}\times\dots\times\mathcal{W}_{N}$.

Throughout this paper, we assume standard properties on the inertia and Coriolis matrices $H_{i}$ and $C_{i}$ that are commonly inherited in most Euler-Lagrange systems  \cite{Ortega-book,Kelly-book}. In particular, we assume the following properties.
\begin{enumerate}[{\bf P1}]
  \item The inertia matrix $H_{i}(q_{i},w_{i})$ is positive definite. More specifically, there are known constants $c_{i,\rm min},c_{i,\rm max}>0$ such that
      \begin{align*}
      c_{i,\rm min}I_{n} \leq H_{i}(q_{i},w_{i}) \leq c_{i,\rm max}I_{n},~~ \forall q_{i}\in\R^{n}, w_{i}\in\mathbb{W}.
      \end{align*}

  \item The matrix-valued function $\dot{H}_{i}(q_{i},\dot{q}_{i},w_{i})-2C_{i}(q_{i},\dot{q}_{i},w_{i})$ is skew symmetric, i.e., for any differentiable function $q_{i}(t)\in\R^{n}$ and its time derivative $\dot{q}_{i}(t)$,
      \begin{equation}\label{Property2}
      \dot{H}_{i}(q_{i},\dot{q}_{i},w_{i}) = C(q_{i},\dot{q}_{i},w_{i}) + C^{\t}(q_{i},\dot{q}_{i},w_{i})
      \end{equation}
      where $\dot{H}_{i}(q_{i},\dot{q}_{i},w_{i}) = \sum_{j=1}^{n}\frac{\partial H_{i}}{\partial q_{ij}}\dot{q}_{ij}$.


  \item The velocity kinematics \eqref{sys-J} linearly depends on a kinematic parameter vector $a(w)\in\R^{p_{i}}$, i.e., there are smooth functions $a_{i}(\cdot)\in\R^{p_{i}}$ and $Z_{i}(\cdot)\in\R^{m \times p_{i}}$ such that for any vectors $q_{i}(t)\in\R^{n}$, $\zeta_{i}(t)\in\R^{m}$,
  \begin{equation}\label{Jlinear}
      J_{g,i}^{T}(q_{i},w_{i})\zeta_{i} = Z_{i}(q_{i},\zeta_{i})a_{i}(w_{i})
  \end{equation}
  where $Z_{i}(\cdot)$ is referred to as a \emph{kinematic regressor matrix} to be known.
  Moreover, there is a smooth matrix-valued function $J_{i}:\R^{n}\times\R^{p_{i}}\to\R^{m\times n}$ such that
  \begin{equation}
    J_{i}(q_{i},a_{i}) = J_{g,i}(q_{i},w_{i}),~~ a_{i} := a_{i}(w_{i})
  \end{equation}
  and consequently $J_{i}^{T}(q_{i},a_{i})\zeta_{i} = J_{g,i}^{T}(q_{i},w_{i})\zeta_{i} = Z_{i}(q_{i},\zeta_{i})a_{i}$.

\end{enumerate}\vspace{0.2cm}

\subsection{Graph on formation}
Let $1 < N  \in\mathbb{N}$ define the number of robotic manipulators whose end-effectors must maintain a specific formation. The neighboring relationships between their end-effectors are described by an undirected and connected graph $\G := \{\V,\E\}$ with the vertex set $\V := \{1,\cdots,N\}$ and the ordered edge set $\E\subset\V\times\V$. The set of the neighbors for the end-effector $i$ is given by $\N_{i} := \{j\in\V:(i,j)\in\E\}$. We use $|\V|=N$ and $|\E|$ to denote the number of vertices and edges of $\G$, respectively. We define the elements of the incidence matrix $B\in\R^{|\V|\times|\E|}$ of $\G$ by
\begin{align*}
b_{ik} = \left\{
\begin{array}{ll} +1, & i=\E_{k}^{\text{tail}} 
\\
-1, &  i=\E_{k}^{\text{head}} 
\\
0, & \text{otherwise} \end{array}
\right.
\end{align*}
where $\E_{k}^{\text{tail}}$ and $\E_{k}^{\text{head}}$ denote the tail and head nodes, respectively, of the edge $\E_{k}$, i.e., $\E_{k}=(\E_{k}^{\text{tail}}, \E_{k}^{\text{head}})$.
Note that $B^{T}\mathbf{1}_{|\V|} = 0$, where $\mathbf{1}_n\in\mathbb{R}^n$ is the vector with all its entries to be ones.

\subsection{End-effector distributed formation control problem}
\begin{figure}
  \centering
  \includegraphics[width=0.36\textwidth]{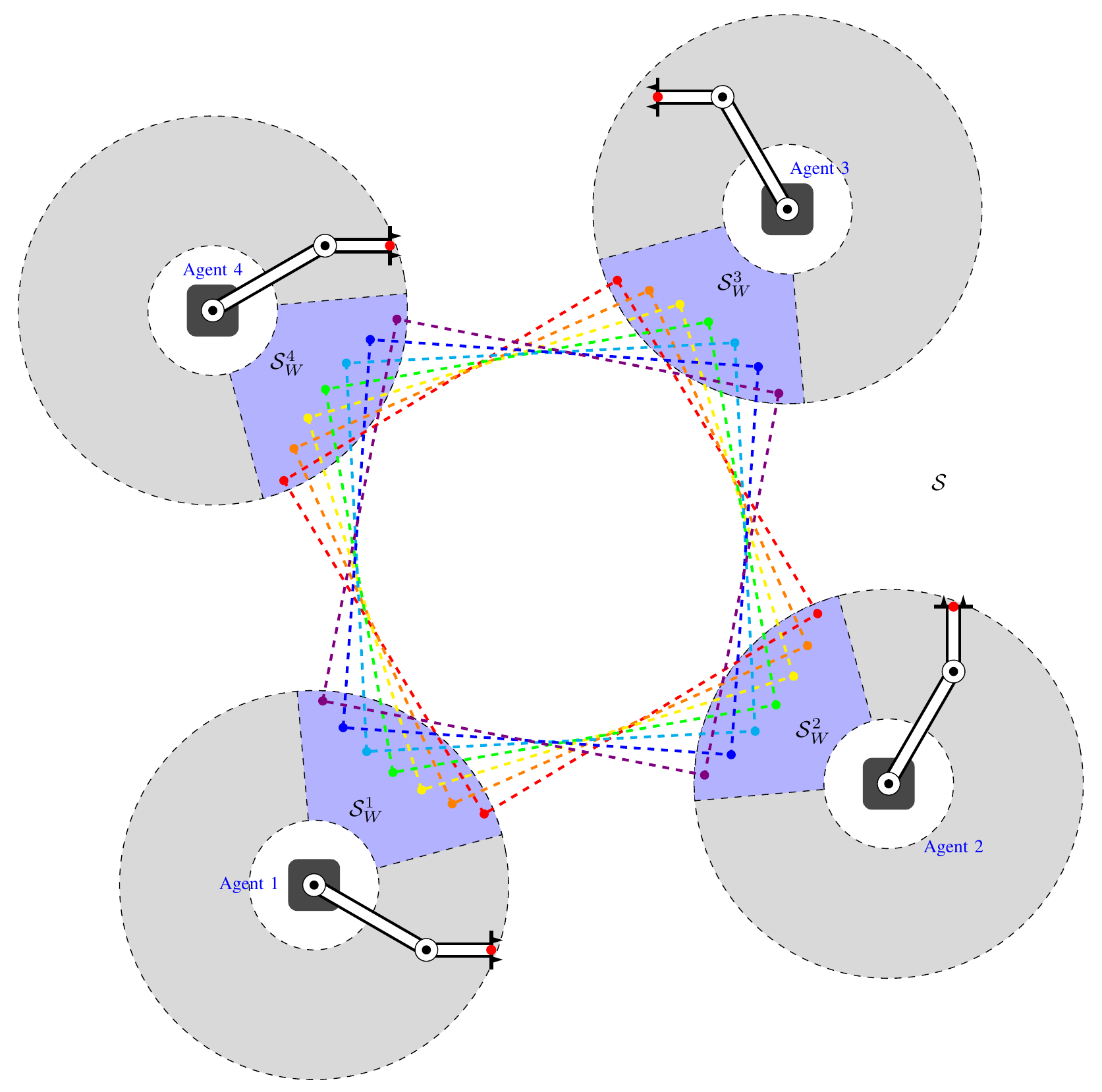}
  \caption{End-effector formation of 4 two-link planar manipulators (whose working space are the gray rings) in the horizontal plane. The desired shape $\mathcal{S}$ ranges in the whole horizontal plane. The reachable desired set $\mathcal{S}_{W} = \mathcal{S}_{W}^{1}\times \mathcal{S}_{W}^{2}\times \mathcal{S}_{W}^{3}\times\mathcal{S}_{W}^{4}$ is the intersection of $\mathcal{S}$ and working space. The dotted squares (in rainbow colors) are possible reachable desired formations. } \label{fig-sets}
\end{figure}

We refer to \emph{configuration} as the stacked vector of end-effectors' positions $x=\col(x_{1},\ldots,x_{N})\in\R^{mN}$, and we refer to \emph{framework} as the pair $(\G,x)$. Given a \emph{reference configuration} $x^*$, we define the \emph{desired shape} as the set
\begin{equation}
    \mathcal{S}:=\{x : x=(I_N \otimes R) x^* + \mathbf{1}_N \otimes b, \, R\in\text{SO($m$)}, b\in\mathbb{R}^m \}.
    \label{eq: ds}
\end{equation}
Let us stack all joint coordinates into $q=\col(q_{1},\ldots,q_{N})$ and $\dot{q} = \col(\dot{q}_{1},\ldots,\dot{q}_{N})$. Note that $\mathcal{S}$ accounts for any arbitrary translation and rotation. 
However, the \emph{working space} for the end-effectors is constrained since we assume that the bases of the arm manipulators are fixed.
Therefore, we define $\mathcal{S}_W = \mathcal{S}\cap\mathcal{W}$ as the subset of shapes that are both desired and reachable by the end-effectors. An illustrative example showing the relationship between $\mathcal{S}$ and $\mathcal{S}_{W}$ is given in Fig.~\ref{fig-sets}.

We are now ready to formulate our formation control problem of end-effectors as follows.
\begin{problem}\label{prob}
\emph{(End-effector distributed formation control problem)}
For a group of $N$ manipulators given by \eqref{sys-dyna}, whose end-effector positions are as in \eqref{sys-kine}, 
design a distributed control law of the form
\begin{equation}
\left\{
\begin{aligned}
\dot{\chi}_{i} &= f_{ci}\Big((x_i - x_j),q_{i},\dot{q}_{i},\chi_{i}\Big) \\
u_{i} &= h_{ci}\Big((x_i - x_j),q_{i},\dot{q}_{i},\chi_{i}\Big)
\end{aligned}
\right. , \quad j\in\mathcal{N}_i
\label{prob-law}
\end{equation}
such that $x(t) \to \mathcal{S}_W$ and $\dot{q}(t) \to \mathbf{0}$ as $t\to\infty$ for 
{\color{black}initial states belong to a region around the desired shape and away from kinematic singularities.}
The state $\chi_{i}(t)$ in (\ref{prob-law}) is the compensator state which will be designed later.
\end{problem}\vspace{0.1cm}

In what follows, we will focus on the distributed control design framework where we can directly extend the well-known distributed formation control of mobile robots (modeled as single-integrator agents) to the formation control of end-effectors in arm manipulators. In the latter case, the dynamics is given by second-order systems as in \eqref{sys-dyna} while the control input is defined at the joint level. In order to illustrate our design framework, we consider the use of displacement-based \cite{de2020maneuvering} and distance-based formation control \cite{de2016distributed}, both of which are fundamental and popular in distributed control methods. We note that our proposed framework is extensible to other gradient-descent based approaches, such as those that are based on the bearing-rigidity framework \cite{zhao2015bearing}.

For the displacement-based formation control, we have that $R = I_m$ in (\ref{eq: ds}). In other words, it only admits desired formation shapes which are given by the translation of $x^*$. On the other hand, the distance-based formation control admits desired formation shapes that are both the translation and rotation of $x^*$. 

The shape displayed by the reference configuration $x^*$ can also be described by a set of geometric relations between the neighboring end-effectors. If $\G$ is connected, then the relative positions defined by the graph $z^* = \overline B^T x^*$ define uniquely the desired shape in displacement-based control, i.e., we have the singleton $\mathcal{Z}_{\text{displacement}} := \{z : z = z^*\}$. Note that the elements of $z^*$ correspond to the ordered $z^*_{ij} = z^*_k := x^*_i - x_j^*, \, (i,j) = \mathcal{E}_k\in\mathcal{E}$. If $\G$ is infinitesimally and minimally rigid (e.g., it has a minimum number of edges for being infinitesimally rigid \cite{AnYuFiHe08}), then the set of distances $\|z_{ij}^*\|, (i,j)\in\mathcal{E}$ define locally\footnote{In the sense that it might define a finite number of other shapes.} the desired shape, i.e., we have the set $\mathcal{Z}_{\text{distance}} := \{z : \|z_{ij}\| = \|x_i - x_j\|, \, (i,j)\in\mathcal{E}\}$.

There are some advantages and disadvantages between the use of displacement-based and distance-based formation control. The former requires a minimum number of edges for $\mathcal{G}$, and the resultant control action for pure kinematic agents is linear. However, the desired shape can only be a translation version of $x^*$, and the algorithm require neighboring agents to share the same frame of coordinates to control the common vector $z_{ij}^*$. On the other hand, the distance-based formation control requires more edges, e.g., at least $(2N - 3)$ in 2D, and the control action for pure kinematic agents is nonlinear leading to only local stability around $\mathcal{S}$. Nevertheless, it allows a more flexible $\mathcal{S}$, e.g., it allows rotations for $x^*$ and the agents do not need to share a common frame of coordinates since they are controlling the scalars $\|z_{ij}^*\|$.

\begin{remark}
Industrial manipulators  use commonly a spherical wrist at the end-effector, and therefore they can achieve any desired orientation at a given end-effector's position \cite[pp.~95]{Murray1994}. This allows us to focus only on the position of the end-effector since its orientation is \emph{decoupled} thanks to the spherical wrist.
\end{remark}

\section{Gradient-based Control Design}\label{sec-grad}

In this section, we propose a distributed control design framework for the case where all the system parameters are exactly known. A preliminary result is also presented on our previous work \cite{Wu2021cdc}.
There are two main elements for each controller $u_{i}$: an end-effector formation controller $u_{i}^{f}$, and a joint velocity controller $u_{i}^{v}$. In Section~\ref{subSec-formation}, we design control law $u_{i}^{f}$ by using virtual spring. Next, in Section~\ref{subSec-joint}, we design $u_{i}^{v}$ based on the passivity property between joint torque and joint velocity. Finally in  Section~\ref{subsec-all}, we present stability and convergence analysis of the overall closed-loop system.

Generally speaking, for solving Problem~\ref{prob}, we firstly employ the virtual spring approach to the end-effectors and introduce standard distributed formation controllers that are based on gradient-descent approach. 
The resulting distributed formation control law defined in the end-effector space is propagated to the joint-space via passivity-based approach.

\subsection{Formation control using virtual spring}\label{subSec-formation}

To achieve the desired formation shape, we start by assigning virtual springs \cite[Chapter~12.2]{van2014pH} on the undirected graph $\G$ of the end-effectors, as depicted in Fig.~\ref{fig-robots}. That is, each edge of $\E$ between the manipulators end-effectors are  interconnected by virtual couplings that shape the energy function of the network. The network's energy function is designed such that its minima are equilibrium points associated to the desired formation shape.

Consider the $k$-th edge between agents $i$ and $j$ connected with a virtual coupling. Let us define the following error signal for each edge $k$ of $\mathcal{G}$
\begin{equation}\label{ek}
e_k(t) := f_e(z_k(t), z_k^*)
\end{equation}
where $f_e : \mathbb{R}^m \to \mathbb{R}^l$, and $l\in\mathbb{N}$ will depend on the chosen formation control strategy, e.g., $f_e = \|z_k\|^2 - \|z_k^*\|^2$  for the  distance-based formation control, and $f_e = z_k - z^*_k$ for the  displacement-based formation control. Each end-effector in the edge $\mathcal{E}_k = (i,j)$ will subsequently use the gradient descent of $V_k(e_k) = \frac{1}{2} \|e_k\|^2$ as its control input (e.g., its velocity when it is described by kinematic point) in order to reach the minimum of $V$ that coincides with the desired shape. It can be checked that the following equality $\nabla_{x_i} V_k = -\nabla_{x_j} V_k \in\mathbb{R}^m$ is satisfied since $z_{ij} = x_i - x_j$. Let us stack all the $e_k$ in $e\in\mathbb{R}^{l|\mathcal{E}|}$ and define $V(e) := \sum_{k=1}^{|\mathcal{E}|}V_k(e_k)$. For compact representation, we define the $m$-dimensional agent-wise displacement measurement $\widehat{e}_{i}(t)\in\R^{m}$, $i=1,\ldots,N$ by
\begin{equation}\label{ei}
\widehat{e}_{i} := \nabla_{x_{i}} V(e)
\end{equation}
or equivalently,
\begin{equation}
\widehat{e}_{i} := \sum_{k=1}^{|\E|} b_{ik} R_{k}(z_{k})e_{k}
\end{equation}
where $R_{k}(z_{k}) = \frac{\partial f_{e}(z_{k},z_{k}^{*})}{\partial z_{k}}$, e.g., for the displacement-based formation control: $R_{k}(z_{k}) = I_{m}$ and for the distance-based formation control: $R_{k}(z_{k}) = 2z_{k}$.

Since the virtual springs are assigned between end-effectors, while the actuators are embedded in joints, the corresponding formation control law $u_{i}^{f}$ of agent $i$ can be written as
\begin{equation}\label{u-i}
u_{i}^{f} = - K_{P}J_{i}^{T}(q_{i},a_{i})\nabla_{x_{i}} V(e) = - K_{P}J_{i}^{T}(q_{i},a_{i})\widehat{e}_{i}
\end{equation}
with design parameter $K_{P}$, where $J_{i}(q_{i},w_{i})$ is the manipulator Jacobian matrix and $\widehat{e}_{i}$ is defined in \eqref{ei}.

Let us stack all the $\widehat{e}_{i}$ in $\widehat{e}\in\R^{mN}$, so that we can write it in the following compact form 
\begin{equation}\label{defn-e-wide}
\widehat{e} = \nabla_x V.
\end{equation}
More precisely, for the displacement-based and for the distance-based ones, we have
\begin{align}
\widehat{e}_\text{displacement} &= \nabla_x V_{\text{displacement-based}} = \overline Be_\text{displacement} \label{eq: disp} \\
\widehat{e}_\text{distance} &= \nabla_x V_{\text{distance-based}} = 2\overline BD_{z}e_\text{distance} \label{eq: dist}
\end{align}
where $D_{z}=\text{block diag}(z_{1},\ldots,z_{|\E|})$.

We note two relevant facts that will be useful for our main technical results. First, $B^TB$ is positive definite if $\mathcal{G}$ does not contain any cycles. Second, $D_{z}^{T}\overline{B^TB}D_{z}$ is positive definite if $\mathcal{G}$ is infinitesimally and minimally rigid. Roughly speaking, infinitesimally rigid means that all the positions $x_i$ are in a generic configuration, e.g., they are not collinear if $m=2$ or coplanar if $m=3$. Note that if the formation is infinitesimally rigid at $\mathcal{S}$, then it is a neighborhood of $\mathcal{S}$ as well.

\subsection{Joint velocity control}\label{subSec-joint}
For solving the static formation control problem, we proceed by designing a control law to stabilize the joint velocity at origin.
Let us define
\begin{equation}\label{xi}
\xi_{i}(t) := \dot{q}_{i}(t)
\end{equation}
which satisfies, by using \eqref{sys-dyna},
\begin{equation}\label{eq: xi}
\dot{\xi}_{i} = H_{i}^{-1}(q_{i},w_{i})\Big(u_{i} - C_{i}(q_{i},\xi_{i},w_{i})\xi_{i} - G_{i}(q_{i},w_{i})\Big).
\end{equation}

According to the well-known passivity of manipulators from joint torque to joint velocity \cite{Spong-book}, we introduce the following controller, consisting of a damping term and a gravity compensation term
\begin{equation}\label{u-v}
u_{i}^{v} = -K_{D}\xi_{i} + G_{i}(q_{i},w_{i})
\end{equation}
with design parameter $K_{D}$.

\newcommand{\nvar}[2]{%
    \newlength{#1}
    \setlength{#1}{#2}
}

\nvar{\dg}{0.3cm}
\def\dw{0.25}\def\dh{0.5}
\nvar{\ddx}{1.5cm}

\def\link{\draw [double distance=1.5mm, very thick] (0,0)--}
\def\joint{%
    \filldraw [fill=white] (0,0) circle (5pt);
    \fill[black] circle (2pt);
}
\def\grip{%
    \draw[ultra thick](0cm,\dg)--(0cm,-\dg);
    \fill (0cm, 0.5\dg)+(0cm,1.5pt) -- +(0.6\dg,0cm) -- +(0pt,-1.5pt);
    \fill (0cm, -0.5\dg)+(0cm,1.5pt) -- +(0.6\dg,0cm) -- +(0pt,-1.5pt);
}
\def\robotbase{%
    \draw[rounded corners=8pt] (-\dw,-\dh)-- (-\dw, 0) --
        (0,\dh)--(\dw,0)--(\dw,-\dh);
    \draw (-0.5,-\dh)-- (0.5,-\dh);
    \fill[pattern=north east lines] (-0.5,-1) rectangle (0.5,-\dh);
}
\def\robotbasenewS{%
    \fill[color = black!72, rounded corners=4pt] (-0.6,-0.6) rectangle (0.6,0.6);
}

\def\camera{%
    \draw[ultra thick](\dg,\dg) rectangle (-0.0cm,-\dg);
    \draw[ultra thick] (0.5\dg, -1.1\dg) -- (\dg, -1.8\dg) -- (0, -1.8\dg) -- (0.5\dg, -1.1\dg);%
    \draw [] (\dg,\dg) node[above]{\footnotesize Camera};
}

\newcommand{\angann}[2]{%
    \begin{scope}[red]
    \draw [dashed, red] (0,0) -- (1.2\ddx,0pt);
    \draw [->, shorten >=3.5pt] (\ddx,0pt) arc (0:#1:\ddx);
    \node at (#1/2-2:\ddx+8pt) {#2};
    \end{scope}
}

\newcommand{\lineann}[4][0.5]{%
    \begin{scope}[rotate=#2, blue,inner sep=2pt]
        \draw[dashed, blue!40] (0,0) -- +(0,#1)
            node [coordinate, near end] (a) {};
        \draw[dashed, blue!40] (#3,0) -- +(0,#1)
            node [coordinate, near end] (b) {};
        \draw[|<->|] (a) -- node[fill=white] {#4} (b);
    \end{scope}
}

\def\thetaone{30}
\def\Lone{2}
\def\thetatwo{-30}
\def\Ltwo{2}
\def\thetathree{-30}
\def\Lthree{2}

\def\thetaoneA{0}
\def\LoneA{2}
\def\thetatwoA{60}
\def\LtwoA{2}
\def\thetathreeA{0}
\def\LthreeA{1}

\def\thetaoneB{120}
\def\LoneB{2}
\def\thetatwoB{60}
\def\LtwoB{2}
\def\thetathreeB{0}
\def\LthreeB{1}

\def\thetaoneC{0}
\def\LoneC{2}
\def\thetatwoC{-60}
\def\LtwoC{2}
\def\thetathreeC{0}
\def\LthreeC{1}

\def\thetaoneD{180}
\def\LoneD{2}
\def\thetatwoD{60}
\def\LtwoD{2}
\def\thetathreeD{0}
\def\LthreeD{1}

\tikzstyle{spring}=[thick,decorate,decoration={aspect=0.5, pre length=0.1cm, post length=0.1cm, segment length=1mm, amplitude=1mm,coil}]

\begin{figure*}[ht]
\centering
\begin{tikzpicture}[scale=0.45]

\robotbasenewS 
\node[mark size=2pt,color=gray!80] (base1) {\pgfuseplotmark{*}};
\draw [] (0,-1) node{\color{black}\footnotesize Agent 1};
\link(\thetaoneA:\LoneA);
\joint
\begin{scope}[shift=(\thetaoneA:\LoneA), rotate=\thetaoneA]
    \link(\thetatwoA:\LtwoA);
    \joint
    \begin{scope}[shift=(\thetatwoA:\LtwoA), rotate=\thetatwoA]
        \grip
        \node[mark size=2pt,color=red] (agent1) {\pgfuseplotmark{*}};
        \begin{scope}[shift=(-\thetatwoA:4), rotate=-\thetatwoA]
            \node[mark size=2pt,color=gray!80] (agent12) {\pgfuseplotmark{*}};
        \end{scope}
        \begin{scope}[shift=(120:1.2), rotate=-120]
            \node[mark size=2pt,color=gray!80] (agent41) {\pgfuseplotmark{*}};
        \end{scope}
    \end{scope}
\end{scope}

\begin{scope}[xshift = 10cm, yshift=1.2cm] 
    \robotbasenewS
    \node[mark size=2pt,color=gray!80] (base2) {\pgfuseplotmark{*}};
    \draw [] (0,-1) node{\color{black}\footnotesize Agent 2};
    \link(\thetaoneB:\LoneB);
    \joint
    \begin{scope}[shift=(\thetaoneB:\LoneB), rotate=\thetaoneB]
        \link(\thetatwoB:\LtwoB);
        \joint
        \begin{scope}[shift=(\thetatwoB:\LtwoB), rotate=\thetatwoB]
            \grip
            \node[mark size=2pt,color=red] (agent2) {\pgfuseplotmark{*}};
       \end{scope}
    \end{scope}
\end{scope}

\begin{scope}[xshift=8.8cm, yshift=8.66cm] 
    \robotbasenewS
    \node[mark size=2pt,color=gray!80] (base3) {\pgfuseplotmark{*}};
    \draw [] (0,-1) node{\color{black}\footnotesize Agent 3};
    \link(\thetaoneD:\LoneD);
    \joint
    \begin{scope}[shift=(\thetaoneD:\LoneD), rotate=\thetaoneD]
        \link(\thetatwoD:\LtwoD);
        \joint
        \begin{scope}[shift=(\thetatwoD:\LtwoD), rotate=\thetatwoD]
            \grip
            \node[mark size=2pt,color=red] (agent3) {\pgfuseplotmark{*}};
            \begin{scope}[shift=(120:1.2), rotate=120]
                \node[mark size=2pt,color=gray!80] (agent23) {\pgfuseplotmark{*}};
            \end{scope}
            \begin{scope}[shift=(-60:4), rotate=-60]
                \node[mark size=2pt,color=gray!80] (agent34) {\pgfuseplotmark{*}};
            \end{scope}
        \end{scope}
    \end{scope}
\end{scope}

\begin{scope}[xshift=-1.2cm, yshift=7.46cm] 
    \robotbasenewS
    \node[mark size=2pt,color=gray!80] (base4) {\pgfuseplotmark{*}};
    \draw [] (0,-1) node{\color{black}\footnotesize Agent 4};
    \link(\thetaoneC:\LoneC);
    \joint
    \begin{scope}[shift=(\thetaoneC:\LoneC), rotate=\thetaoneC]
        \link(\thetatwoC:\LtwoC);
        \joint
        \begin{scope}[shift=(\thetatwoC:\LtwoC), rotate=\thetatwoC]
            \grip
            \node[mark size=2pt,color=red] (agent4) {\pgfuseplotmark{*}};
        \end{scope}
    \end{scope}
\end{scope}

\draw [spring,gray!40] (agent1)  -- node[below] {\color{gray!60}$z_{1X}$} (agent12);
\draw [spring,gray!40] (agent12) -- node[left] {\color{gray!60}$z_{1Y}$} (agent2);
\draw [spring,gray!40] (agent2)  -- node[right] {\color{gray!60}$z_{2Y}$} (agent23);
\draw [spring,gray!40] (agent23) -- node[below] {\color{gray!60}$z_{2X}$} (agent3);
\draw [spring,gray!40] (agent3)  -- node[above left] {\color{gray!60}$z_{3X}$} (agent34);
\draw [spring,gray!40] (agent34) -- node[right] {\color{gray!60}$z_{3Y}$} (agent4);
\draw [spring,gray!40] (agent4)  -- node[left] {\color{gray!60}$z_{4Y}$} (agent41);
\draw [spring,gray!40] (agent41) -- node[above] {\color{gray!60}$z_{4X}$} (agent1);

\draw[-latex,thick,red] ($(base1)+(0,0)$) -- ($(base1)+(1.8,0)$);
\draw[-latex,thick,red] ($(base1)+(0,0)$) node[right, yshift=0.5cm] {\footnotesize\color{red}$\Sigma_{1}$} -- ($(base1)+(0,1.8)$);
\draw[-latex,thick,red] ($(base2)+(0,0)$) -- ($(base2)+(1.8,0)$);
\draw[-latex,thick,red] ($(base2)+(0,0)$) node[right, yshift=0.5cm] {\footnotesize\color{red}$\Sigma_{2}$} -- ($(base2)+(0,1.8)$);
\draw[-latex,thick,red] ($(base3)+(0,0)$) -- ($(base3)+(1.8,0)$);
\draw[-latex,thick,red] ($(base3)+(0,0)$) node[right, yshift=0.5cm] {\footnotesize\color{red}$\Sigma_{3}$} -- ($(base3)+(0,1.8)$);
\draw[-latex,thick,red] ($(base4)+(0,0)$) -- ($(base4)+(1.8,0)$);
\draw[-latex,thick,red] ($(base4)+(0,0)$) node[right, yshift=0.5cm] {\footnotesize\color{red}$\Sigma_{4}$} -- ($(base4)+(0,1.8)$);


\draw []  (5,-1.8) node{\small Displacement-based formation setup};

\begin{scope}[xshift=-3cm, yshift=3cm] 
    \node[mark size=2pt,color=red] (world) {\pgfuseplotmark{*}};
\draw[-latex,thick,red] ($(world)+(0,0)$) node[left] {\footnotesize\color{red}$\Sigma_{g}$} -- ($(world)+(1.8,0)$) node[below] {\footnotesize\color{red}$X$};
\draw[-latex,thick,red] ($(world)+(0,0)$) -- ($(world)+(0,1.8)$) node[left] {\footnotesize\color{red}$Y$} ;
\end{scope}

\begin{scope}[xshift = 18cm, yshift=0cm] 
\robotbasenewS 
\node[mark size=2pt,color=gray!80] (base1) {\pgfuseplotmark{*}};
\draw [] (0,-1) node{\color{black}\footnotesize Agent 1};
\link(\thetaoneA:\LoneA);
\joint
\begin{scope}[shift=(\thetaoneA:\LoneA), rotate=\thetaoneA]
    \link(\thetatwoA:\LtwoA);
    \joint
    \begin{scope}[shift=(\thetatwoA:\LtwoA), rotate=\thetatwoA]
        \grip
        \node[mark size=2pt,color=red] (agent1) {\pgfuseplotmark{*}};
    \end{scope}
\end{scope}

\begin{scope}[xshift = 10cm, yshift=1.2cm] 
    \robotbasenewS
    \node[mark size=2pt,color=gray!80] (base2) {\pgfuseplotmark{*}};
    \draw [] (0,-1) node{\color{black}\footnotesize Agent 2};
    \link(\thetaoneB:\LoneB);
    \joint
    \begin{scope}[shift=(\thetaoneB:\LoneB), rotate=\thetaoneB]
        \link(\thetatwoB:\LtwoB);
        \joint
        \begin{scope}[shift=(\thetatwoB:\LtwoB), rotate=\thetatwoB]
            \grip
            \node[mark size=2pt,color=red] (agent2) {\pgfuseplotmark{*}};
       \end{scope}
    \end{scope}
\end{scope}

\begin{scope}[xshift=8.8cm, yshift=8.66cm] 
    \robotbasenewS
    \node[mark size=2pt,color=gray!80] (base3) {\pgfuseplotmark{*}};
    \draw [] (0,-1) node{\color{black}\footnotesize Agent 3};
    \link(\thetaoneD:\LoneD);
    \joint
    \begin{scope}[shift=(\thetaoneD:\LoneD), rotate=\thetaoneD]
        \link(\thetatwoD:\LtwoD);
        \joint
        \begin{scope}[shift=(\thetatwoD:\LtwoD), rotate=\thetatwoD]
            \grip
            \node[mark size=2pt,color=red] (agent3) {\pgfuseplotmark{*}};
        \end{scope}
    \end{scope}
\end{scope}

\begin{scope}[xshift=-1.2cm, yshift=7.46cm] 
    \robotbasenewS
    \node[mark size=2pt,color=gray!80] (base4) {\pgfuseplotmark{*}};
    \draw [] (0,-1) node{\color{black}\footnotesize Agent 4};
    \link(\thetaoneC:\LoneC);
    \joint
    \begin{scope}[shift=(\thetaoneC:\LoneC), rotate=\thetaoneC]
        \link(\thetatwoC:\LtwoC);
        \joint
        \begin{scope}[shift=(\thetatwoC:\LtwoC), rotate=\thetatwoC]
            \grip
            \node[mark size=2pt,color=red] (agent4) {\pgfuseplotmark{*}};
        \end{scope}
    \end{scope}
\end{scope}

\draw [spring,gray!40] (agent1) -- node[below] {\color{gray}$z_{1}$} (agent2);
\draw [spring,gray!40] (agent2) -- node[above right] {\color{gray}$z_{2}$} (agent3);
\draw [spring,gray!40] (agent3) -- node[above left] {\color{gray}$z_{3}$} (agent4);
\draw [spring,gray!40] (agent4) -- node[above left] {\color{gray}$z_{4}$} (agent1);
\draw [spring,gray!40] (agent1) -- node[above left] {\color{gray}$z_{5}$} (agent3);

\draw[-latex,thick,red] ($(base1)+(0,0)$) -- ($(base1)+(1.8,0)$);
\draw[-latex,thick,red] ($(base1)+(0,0)$) node[right, yshift=0.5cm] {\footnotesize\color{red}$\Sigma_{1}$} -- ($(base1)+(0,1.8)$);
\draw[-latex,thick,red] ($(base2)+(0,0)$) -- ($(base2)+(1.8,0)$);
\draw[-latex,thick,red] ($(base2)+(0,0)$) node[right, yshift=0.5cm] {\footnotesize\color{red}$\Sigma_{2}$} -- ($(base2)+(0,1.8)$);
\draw[-latex,thick,red] ($(base3)+(0,0)$) -- ($(base3)+(1.8,0)$);
\draw[-latex,thick,red] ($(base3)+(0,0)$) node[right, yshift=0.5cm] {\footnotesize\color{red}$\Sigma_{3}$} -- ($(base3)+(0,1.8)$);
\draw[-latex,thick,red] ($(base4)+(0,0)$) -- ($(base4)+(1.8,0)$);
\draw[-latex,thick,red] ($(base4)+(0,0)$) node[right, yshift=0.5cm] {\footnotesize\color{red}$\Sigma_{4}$} -- ($(base4)+(0,1.8)$);


\draw []  (5,-1.8) node{\small Distance-based formation setup};

\end{scope}

\end{tikzpicture}
\caption{End-effector formation of 4 two-link planar manipulators in  the horizontal plane. Left: displacement-based formation control. Right: distance-based formation control. The springs in gray color are virtual couplings between end-effectors. $
\Sigma_{1}$ to $\Sigma_{4}$ in red color are frames attached at manipulator bases.}\label{fig-robots}
\end{figure*}
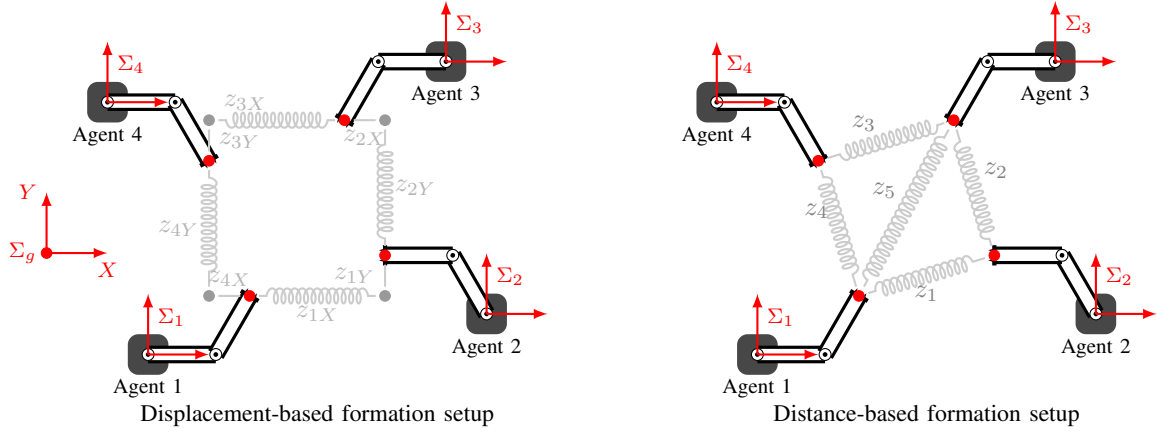

\subsection{Closed-loop system}\label{subsec-all}
In this part, we will combine the individual control laws $u_{i}^{f}$ and $u_{i}^{v}$ above and analyze the solvability of Problem \ref{prob}.

Before presenting the following main result, for the rest of the paper and for the sake of presentation convenience, we denote: $c_{\min} = \min_{i\in\{1,\ldots,N\}}\{c_{i,\min}\}$, $c_{\max} = \max_{i\in\{1,\ldots,N\}}\{c_{i,\max}\}$,
$q=\col(q_{1},\dots,q_{N})$, $\xi=\col(\xi_{1},\dots,\xi_{N})$, $w=\col(w_{1},\dots,w_{N})$, $\widehat{e} = \col(\widehat{e}_{1},\dots,\widehat{e}_{N})$, $e=\col(e_{1},\dots,e_{|\E|})$, $x_{0}=\col(x_{10},\dots,x_{N0})$, $h(q,w)=\col(h_{1}(q_{1},w_{1}),\dots,h_{N}(q_{N},w_{N}))$,
$H(q,w)=\textnormal{block diag}(H_{1}(q_{1},w_{1}),\dots,H_{N}(q_{N},w_{N}))$, $C(q,\xi,w)=\textnormal{block diag}(C_{1}(q_{1},\xi_{1},w_{1}),\dots,C_{N}(q_{N},\xi_{N},w_{N}))$, $G(q,w)=\textnormal{block diag}(G_{1}(q_{1},w_{1}),\dots,G_{N}(q_{N},w_{N}))$, $J(q,w)=\textnormal{block diag}(J_{1}(q_{1},w_{1}),\dots,J_{N}(q_{N},w_{N}))$. Finally, we denote $\mathrm{Q}:=\{q\in\R^{nN} : \text{det}(J(q,w)J^{T}(q,w))=0\}$.

\smallskip
\begin{theorem}\label{thm-I}
Consider $N$ robot manipulators \eqref{sys-dyna} satisfying assumptions \textbf{P1} and \textbf{P2} where the system parameters are perfectly known. Assume that the formation graph is  infinitesimally and minimally rigid graph $\G$.  
Then for any end-effector reference configuration $x^*$,
the end-effector formation control problem can be solved by the following distributed control law, for $i=1,\ldots,N$,
\begin{align}\label{law-I}
u_{i} &= - K_{P}J_{i}^{T}(q_{i},a_{i}) \widehat{e}_{i} - K_{D}\xi_{i} + G_{i}(q_{i},w_{i})
\end{align}
with sufficiently large positive gains $K_{P}$ and $K_{D}$, where $\widehat{e}_{i}$, $\xi_{i}$ are given in \eqref{ei}, \eqref{xi}, respectively. \\
Particularly, there exist constants $K_{P,\text{min}},K_{D,\text{min}},\alpha,r,\epsilon>0$ for the closed-loop system \eqref{sys-dyna}, \eqref{law-I} such that if $K_P>K_{P,\text{min}}$, $K_D>K_{D,\text{min}}$, $(e(0),\xi(0))\in \mathcal{S}_{W_{r}} := \{e,\xi: \frac{K_{P}+2\alpha K_{D}}{4}\|e\|^{2} + \frac{c_{\min}+c_{\max}}{4}\|\xi\|^{2} \leq r\}$ and 
$q(0)\in\mathbb{Q}_{\epsilon} := \{q : \text{dist} (q,\mathrm{Q}) > \epsilon \}$ then the origin $(e,\xi) = (\mathbf{0}, \mathbf{0})$ of the error dynamics is exponentially stable.
\end{theorem}\vspace{0.2cm}

\begin{remark}
The set $\mathcal{S}_{W_{r}}$ describes how ``close'' (quantified by $r$) the manipulators are to the desired shape, and the set $\mathbb{Q}_{\epsilon}$ describes how ``far'' (quantified by $\epsilon$) the manipulators are from the kinematic singularities.
The former directly implies that our result is only valid in a local sense. The latter is a common condition for manipulators to operate away from singular configurations.
Note that even for two configurations with the same $r$, their $\epsilon$ can be different. An illustrative example is shown in Figure \ref{fig-sets}, where both desired squares red and light green have $r=0$ in $\mathcal{S}_{W_{r}}$; however, the former admits a smaller $\epsilon$ than the latter. This is because the red square is closer to singular configurations, where the Jacobian is not full rank as opposed to that of the green one.
\end{remark}


It is worth noting that the proposed design of distributed control protocols in Theorem \ref{thm-I} above 
is applicable to both displacement-based and  distance-based formation control by defining appropriately the potential function of the formation $V$ in \eqref{ei} and \eqref{law-I}. In this paper, we only show the proof for the distance-based case. The proof for the displacement-based case can be obtained following the same procedure and is omitted here.


\textit{Proof of Theorem~\ref{thm-I}:}
The proof is routine in Lyapunov's direct method.
Substituting control law \eqref{law-I} into 
\eqref{sys-dyna} and using 
\eqref{ek}, 
the closed-loop system can compactly be written as
\begin{equation}\label{cls-I}
\left\{
\begin{aligned}
\dot{e} &= 2D_{z}^{T}\overline{B}^{T}J(q,a)\xi \\
\dot{\xi} &= H^{-1}(q,w)[-K_{P}J^{T}(q,a)\widehat{e} - K_{D}\xi - C(q,\xi,w)\xi]
\end{aligned}
\right.
\end{equation}
which is a nonautonomous system because the singularity-free $q:=q(t)$ is considered here as a time-varying exosignal satisfying $\dot{q} = \xi$ and $e = \col(\dots,\|h_{i}(q_{i},w_{i}) + x_{i0} - h_{j}(q_{j},w_{j}) - x_{j0} \|^{2} - \|z_{k}^*\|……{2},\dots)$. 

Let $\mathcal{Q}$ and $\Xi$ be the sets $\{e\in\R^{l |\E|} : \|e\|^{2}\leq r_{1}\}$ and $\{\xi\in\R^{nN} : \|\xi\|^{2} \leq r_{2}\}$ for some $r_{1},r_{2}>0$, respectively.
Let us define a Lyapunov function candidate $U_{1}:[0,\infty)\times\mathcal{Q}\times\Xi\to\R$  by
\begin{align}\label{U1}
U_{1}(t,e,\xi)
&= \frac{1}{4}e^{T}(K_{P} + \alpha K_{D})e + \frac{1}{2}\xi^{T}H(q(t),w)\xi \nnum\\
&\quad + \alpha \widehat{e}^{T}J(q(t),a)H(q(t),w)\xi
\end{align}
where $\alpha$ is any constant satisfying $0 < \alpha < \frac{c_{\min}}{c_{\max}}$ with $c_{\min} = \min_{i\in\{1,\ldots,N\}}\{c_{i,\min}\}$, $c_{\max} = \max_{i\in\{1,\ldots,N\}}\{c_{i,\max}\}$, and $c_{i,\min}$, $c_{i,\max}$, $i\in\{1,\ldots,N\}$ be as in {\bf P1}.

The remaining proof is divided into three parts as follows.

{\sl Part~1.}
Let us show that $U_{1}$ is a positive definite function on $\mathcal{Q}\times\Xi$ (uniformly on the exosignal $q(t)$ satisfying the joint displacement constraints).
By using Young's inequality, the cross term of \eqref{U1} satisfies
\begin{align}\label{U1-cross}
&\alpha\widehat{e}^{T}J(q,a)H(q,w)\xi \nnum\\
&\leq  \alpha\|\widehat{e}^{T}J(q,a)\| \|H(q,w)\| \|\xi\|   \nnum\\
&\leq \frac{1}{2}\alpha\|H(q,w)\| \Big(\|\widehat{e}^{T}J(q,a)\|^{2} + \|\xi\|^{2}\Big).
\end{align}
Using $\widehat{e}=2D_{z}^{T}\overline{B}^{T}e$, we have
\begin{equation}\label{lambda1}
\|\widehat{e}^{T}J(q,a)\|^{2} \leq \|J(q,a)\|^{2} \|2D_{z}^{T}\overline{B}^{T}e\|^{2} \leq \lambda_{1}\lambda_{J} \|e\|^{2}
\end{equation}
where $\lambda_{1} := \lambda_{\max}\{4D_{z}^{T}\overline{B}^{T}\overline{B}D_{z}\}$ for all $e\in\mathcal{Q}$ since $z_l^T z_n, (l,n)\in\mathcal{E}$ can be written as a functions of $e$ \cite{mou2015undirected}, and $\lambda_{J} = \lambda_{\max}\{J(q,a)J^{T}(q,a)\}$ for all admissible $q$ and $w\in\mathbb{W}$.
Then, using \eqref{U1-cross}, \eqref{lambda1} and {\bf P1}, it follows that
\begin{equation}\label{U1-1}
\frac{1}{2}c_{01}\|e\|^{2} + \frac{1}{2}c_{02}\|\xi\|^{2} \leq U_{1} \leq \frac{1}{2}c_{03}\|e\|^{2} + \frac{1}{2}c_{04}\|\xi\|^{2}
\end{equation}
for all admissible $q(t)$, $(e,\xi)\in\mathcal{Q}\times\Xi$,
where
\begin{equation}\label{U1-1-c}
\begin{aligned}
c_{01} &= \frac{1}{2}(K_{P} + \alpha K_{D}) - \alpha c_{\max}\lambda_{1}\lambda_{J} \\
c_{02} &= c_{\min} - \alpha c_{\max}  \\
c_{03} &= \frac{1}{2}(K_{P} + \alpha K_{D}) + \alpha c_{\max}\lambda_{1}\lambda_{J} \\
c_{04} &= c_{\max} + \alpha c_{\max}.
\end{aligned}
\end{equation}
Hence, for any $0 <\alpha < \frac{c_{\min}}{c_{\max}}$, $U_{1}$ is positive definite if the design parameters are chosen to satisfy
\begin{equation}\label{U1-1-c}
K_{P} > 0, ~~ K_{D} > 2c_{\max}\lambda_{1}\lambda_{J}.
\end{equation}

\vspace{0.2cm}
{\sl Part~2.} This part is to show that the time derivative of \eqref{U1} is negative definite (uniformly with respect to the admissible $q(t)$).
By calculating the time derivative of \eqref{U1}, we obtain that 
\begin{align}\label{U1-t}
& \dot{U}_{1}|_{\eqref{cls-I}} = e^{T}(K_{P}+\alpha K_{D})D_{z}^{T}\overline{B}^{T}J(q(t),a)\xi \nnum\\
&\quad + \xi^{T}\Big[-K_{P}J^{T}(q(t),a)\widehat{e} - K_{D}\xi - C(q(t),\xi,w)\xi\Big] \nnum\\
&\quad + \frac{1}{2}\xi^{T}\dot{H}(q(t),w)\xi + \lambda \dot{\widehat{e}}^{T}J(q,a)H(q(t),w)\xi \nnum\\
&\quad  + \alpha \widehat{e}^{T}J(q(t),a)\Big[-K_{P}J^{T}(q(t),a)\widehat{e}  - K_{D}\xi  \nnum \\
& \qquad - C(q(t),\xi,w)\xi \Big] 
+ \alpha \widehat{e}^{T}\dot{J}(q(t),a)H(q(t),w)\xi \nnum \\
& \qquad + \alpha \widehat{e}J(q(t),a)\dot{H}(q(t),w)\xi
\end{align}
where
\begin{gather}
\dot{\widehat{e}} = 2\overline{B}D_{(\overline{B}^{T}J(q(t),a)\xi)}e + 4\overline{B}D_{z} D_{z}^{T}\overline{B}^{T}J(q(t),a)\xi \nnum\\
\dot{J}(q(t),a) = \sum_{i=1}^{N}\sum_{j=1}^{n} \frac{\partial J}{\partial q_{ij}}\xi_{ij} \label{J-dt1} \\
\dot{H}(q(t),w) = C(q(t),\xi,w) + C^{T}(q(t),\xi,w) \nnum.
\end{gather}
Note that the last equality is due to property {\bf P2}.
Using \eqref{eq: dist} and removing common terms in \eqref{U1-t}, we can rewrite \eqref{U1-t} as
\begin{align}\label{U1-d}
\dot{U}_{1}|_{\eqref{cls-I}} &= - \alpha \widehat{e}^{T}J(q(t),a)K_{P}J^{T}(q(t),a)\widehat{e} - \xi^{T}K_{D}\xi \nnum\\
&\quad + \alpha \phi_{1}(t,e,\xi,w)
\end{align}
where
\begin{align}\label{phi-1}
 \phi_{1}(t,e,\xi,w) &= \dot{\widehat{e}}^{T}J(q(t),a)H(q(t),w)\xi \nnum \\ & \qquad +  \widehat{e}^{T}\dot{J}(q(t),a)H(q(t),w)\xi \nnum\\
&\qquad
+  \widehat{e}^{T}J(q(t),a)C^{T}(q(t),\xi,w)\xi.
\end{align}

Following the time derivative in \eqref{U1-d}, we can state the following.
\begin{itemize}
    \item It can be shown that $\phi_{1}(t,e,\xi,w)$ is a smooth function satisfying $\phi_{1}(t,0,0,w) = 0$ for all admissible $q(t)$ and all $w\in\mathbb{W}$. As shown in the Appendix, there are positive constants $k_{11}$ and $k_{12}$ such that
    \begin{equation}\label{phi1-Young}
    \|\phi_{1}(t,e,\xi,w)\| \leq k_{11}\|e\|^{2}  + k_{12}\|\xi\|^{2}
    \end{equation}
    for all $t$, all $e\in\mathcal{Q}$, all $\xi\in\Xi$, and all $w\in\mathbb{W}$.

    \item In the distance-based formation control, we note that the elements of  $D_{z}^{T}\overline{B^{T}B}D_{z}$ are of the form $z_{i}^{T}z_{j}$, $i,j=1,\ldots,|\E|$. It allows us to write $Q(e):=D_{z}^{T}\overline{B^{T}B}D_{z}$ as in \cite{Marina2018taming}. Since $\G$ is infinitesimally and minimally rigid (as assumed in the theorem), $Q(0) = D^{T}_{z^{*}}\overline{B^{T}B}D_{z^{*}} $ is positive definite. Then since the eigenvalues of a matrix are continuous functions of their entries, we have $Q(e)$ is positive definite in the set $\mathcal{Q}$ for some $r_{1}\geq 0$. Since the manipulators operate away from kinematics singularity set $Q$, we have that $\overline{B}D_{z}J(q(t),a)J^{T}(q(t),a)D_{z}^{T}\overline{B}^{T}$ is positive definite in the set $\mathcal{Q}$.
\end{itemize}

By substituting $\widehat{e} = 2\overline{B}D_{z}e$ and \eqref{phi-1} into \eqref{U1-d}, we have
\begin{align*}
\dot{U}_{1}|_{\eqref{cls-I}} &\leq -\alpha(\lambda_{2} K_{P} - k_{11})\|e\|^{2} - (K_{D} - k_{12})\|\xi\|^{2}
\end{align*}
where $\lambda_{2} := \lambda_{\min}\{D_{z}^{T}\overline{B}^{T}J(q,a)J^{T}(q,a)\overline{B}D_{z}\}$ for all admissible $q$ and $e\in\mathcal{Q}$. Note that the minimum eigenvalue is bigger than zero since the formation is minimally and infinitesimally rigid, and the Jacobian is not singular.
The above time derivative can be made negative by the following steps:
\begin{enumerate}[1)]
    \item  Fix the constant $\alpha$ such that $0 < \alpha < \frac{c_{\min}}{c_{\max}}$.
    \item Compute $k_{11}$ and $k_{12}$ of \eqref{phi1-Young} for some $r_{1},r_{2}>0$.
    \item Compute $\lambda_{2}$ for the same $r_{1}$.
    \item Choose $K_{P}$ and $K_{D}$ such that
    \begin{equation}\label{U1-2-c}
    \lambda_{2} K_{P} - k_{11} >1,~~
    K_{D} - \alpha k_{12} > 1.
    \end{equation}
\end{enumerate}

Then, we obtain
\begin{equation}\label{U1-2}
\dot{U}_{1}|_{\eqref{cls-I}} \leq - \|e\|^{2} - \|\xi\|^{2}
\end{equation}
for all admissible $q(t)$, all $(e,\xi)\in\mathcal{Q}\times\Xi$, and all $w\in\mathbb{W}$.
Therefore, we have $K_{P,\min} = \frac{1}{\lambda_{2}}(k_{11} + 1)$ and $K_{D,\min} = \{\alpha k_{12}+1, 2c_{\max}\lambda_{1}\lambda_{J}\}$ from \eqref{U1-1-c} and \eqref{U1-2-c}.

\vspace{0.2cm}
{\sl Part~3.} Note that the previous two steps are on the time-varying Lyapunov function $U_1$ that depends on the joint's velocities $\xi$, and the \emph{distortion} of the shape measured by the error signal $e$. In this third part, we 
will show that the final positions of the end-effectors converge to $\mathcal{S}_W$.
Recall the bounds of $U_1$ in \eqref{U1-1} and let us define the following sets
\begin{align*}
\Omega_{1,r} &= \{(e,\xi)\in\R^{mN}\times\R^{nN}: c_{01}\|e\|^{2} + c_{02}\|\xi\|^{2} \leq c_{01}r\} \\
\Omega_{2,r} &= \{(e,\xi)\in\R^{mN}\times\R^{nN}: c_{03}\|e\|^{2} + c_{04}\|\xi\|^{2} \leq c_{01}r\}
\end{align*}
for some $r>0$.
It can be verified that $\mathcal{S}_{W_{r}} \subset \Omega_{2,r} \subset \Omega_{1,r}$, and $r$ can be chosen such that $\Omega_{1,r}\subset\mathcal{Q}\times\Xi$.



Since $\dot{U}_{1}|_{\eqref{cls-I}} \leq 0 $, $U_{1}(t,e(t),\xi(t)) \leq U_{1}(0,e(0),\xi(0))$ for all $t\geq 0$. It follows that any solution starting in $\mathcal{S}_{W_{r}}$ stays in $\Omega_{1,r}$, and consequently in $\mathcal{Q}\times\Xi$ for all $t$. Hence, the solution is bounded for all $t$.
Moreover, by combining \eqref{U1-1}, \eqref{U1-2} and using \cite[Theorem~4.10]{Khalil2002}, the origin defined by the signals $e(t)$ and $\xi(t)$ is exponentially stable.
Then, we have that the joint's velocities $\xi(t)\to\mathbf{0}$ exponentially fast as $t\to\infty$; therefore, the total distance travelled by the end-effectors is bounded. Hence, if $q(0)\in\mathbb{Q}_{\epsilon}$ with $\epsilon$ sufficiently large, $x(t)\to\mathcal{S}_{W}$ as $t\to\infty$ with $q(t)$ be always away from the kinematics singularities. 

\hfill Q.E.D.




\section{Robust control redesign with respect to parameters uncertainties}\label{sec-robust}

The control law \eqref{law-I} proposed in the previous section requires complete knowledge on  
system parameters. Specifically, $u_{i}^{v}$ of \eqref{u-v} needs information on  parameters for exact gravitational compensation, and $u_{i}^{f}$ of \eqref{u-i} needs kinematic parameters in the Jacobian matrix for stabilization control. This knowledge requirement limits the robustness of the resulting closed-loop system. Although robust control of single manipulator's end-effector has been studied in literature (e.g., \cite{Cheah1999pid,Cheah2003approximate,Dixon2007adaptive}), imprecision in parameters remains an issue if the task has to be done in a distributed way by a team of manipulators, i.e., 
central monitoring and control is not allowed. 

Correspondingly, we investigate this particular problem in this section, where as before 
a team of manipulators, whose dynamic and kinematic parameters are not exactly known, has to solve Problem \ref{prob}. Without loss of generality, we can assume that the parameter vector $w_{i}$ of \eqref{sys-dyna} is written in the form
\begin{equation}
w_{i} = \hat{w}_{i} + \Delta w_{i}
\end{equation}
where $\hat{w}_{i}$ represents the nominal part (or approximate value) while $\Delta w_{i}$ represents the uncertain part. 
In this scenario, a direct application of \eqref{law-I} is to use $\hat{w}_{i}$ instead of $w_{i}$, e.g.
\begin{align}\label{law-I-est}
u_{i} &= - K_{P}J_{i}^{T}(q_{i},\hat{a}_{i}) \widehat{e}_{i} - K_{D}\xi_{i} + G_{i}(q_{i},\hat{w}_{i})
\end{align}
where $\hat{a}_{i} := a_{i}(\hat{w}_{i})$. However, this could lead to the following two immediate consequences. Firstly, the equilibrium point of the closed-loop system \eqref{sys-dyna} and \eqref{law-I-est} at the origin can be shifted if $G_{i}(q_{i},\hat{w}_{i}) \neq G_{i}(q_{i},w_{i})$ at the desired shape, i.e., $e$ might tend to a non-zero constant vector. Secondly, the mismatch between the nominal Jacobian matrix $J_{i}(q_{i},\hat{a}_{i})$ and the actual Jacobian matrix $J_{i}(q_{i},a_{i})$ may destabilize the closed-loop system.

To overcome these drawbacks, we will modify \eqref{law-I-est} such that it can accommodate for parametric uncertainties. Section~\ref{sec-im} presents an additional dynamic compensator for the gravitation compensation. Section~\ref{sec-st} handles the kinematics uncertainties in the Jacobian matrix.

\subsection{Dynamic compensator design}\label{sec-im}
For a given desired shape, the manipulators have a desired joint-space configuration given by  
$q^{*}\in\{ q\in\mathbb{R}^{nN} : h(q,w) + x_{0}\in\mathcal{S}_{W} \}$. 
In the case where $G_{i}(q_{i}^{*},\hat{w}_{i}) \neq G_{i}(q_{i}^{*},w_{i})$ and $G_{i}(q_{i}^{*},w_{i})\neq \mathbf{0}$, asymptotic convergence to the desired shape can not be achieved by \eqref{law-I-est} due to the lack of 
steady-state error compensation. Hence, an additional compensator is required for asymptotic convergence.

The design of dynamic compensator is based on the internal model principle \cite[Chapter~5]{Khalil-book}, which requires the use of integral action to ensure zero steady-state error in the presence of parameter uncertainties.
To compensate for gravity, we introduce the following dynamics
\begin{equation}\label{im}
    \dot{\eta}_{i} = - K_{I}\eta_{i} + u_{i}
\end{equation}
where $K_{I}$ is a positive constant.
Let $\eta^{*}$ be the steady-state of the stacked vector $\eta = \col(\eta_{1},\ldots,\eta_{N})$. It can be verified that $\eta^{*} = K_{I}^{-1}G(q^{*},w)$.
By defining 
coordinate transformation
 $\tilde{\eta} = \eta - \eta^{*} - H(q,w)\xi$,
it follows immediately that we have 
the following error dynamics
\begin{align}\label{im-e}
    \dot{\tilde{\eta}} 
    &= - K_{I}\tilde{\eta} - K_{I}H(q,w)\xi - C^{T}(q,\xi,w)\xi  \nnum\\
    &\quad + G(q,w) - G(q^{*},w).
\end{align}
Using storage function
\begin{equation}\label{V-eta}
V_{\eta}(\tilde{\eta}) = \frac{1}{2}\tilde{\eta}^{T}K_{I}^{-1}\tilde{\eta}
\end{equation}
there are constants $k_{21},k_{22}>0$ such that
\begin{align}\label{V-eta-d}
\dot{V}_{\eta}|_{\eqref{im-e}} &= -\tilde{\eta}^{T}\tilde{\eta} + \tilde{\eta}^{T}\Big[- H(q,w)\xi - K_{I}^{-1}C^{T}(q,\xi,w)\xi \nnum\\
&\quad + K_{I}^{-1}G(q,w) - K_{I}^{-1}G(q^{*},w)\Big] \nnum\\
&\leq -\frac{1}{2} \|\tilde{\eta}\|^{2} + k_{21}\|e\|^{2} + k_{22}\|\xi\|^{2}
\end{align}
for all $t\geq0$, all $\tilde{\eta}\in\R^{nN}$, all $(e,\xi)\in\mathcal{Q}\times\Xi$ and all $w\in\mathbb{W}$. Explicit expressions of $k_{21}$ and $k_{22}$ is put in Appendix.

\subsection{Robust stabilization}\label{sec-st}

In the first part of this subsection, we analyze the asymptotic stability of the closed-loop systems when we can only rely on a limited information about the nominal (approximate) Jacobian matrix and the bound of the mismatches with respect to the actual ones. When the bound is sufficiently small, we present a sufficient condition on the control gains that guarantees asymptotic stability.
In the second part, we propose an adaptive Jacobian control law that can relax the above mentioned mismatch bound. The adaptive law uses direct cancellation and guarantees asymptotic stability of the closed-loop systems.




\medskip
\subsubsection{Approximate Jacobian approach}\label{sec-J-I}

Let us assume the following property on the Jacobian matrix. 

\begin{itemize}
    \item [{\bf P4}] The mismatch between the real Jacobian matrix $J(q,a)$ and the nominal Jacobian matrix $J(q,\hat{a})$ is upper-bounded in the following sense: 
    there is a known positive constant $\delta$ such that
    \begin{equation}\label{delta}
    \|J(q, \tilde{a}) \| \leq \delta,~~  J(q, \tilde{a}) :=  J(q,\hat{a}) - J(q,a)
    \end{equation}
    holds for all admissible $q$ and all $w\in\mathbb{W}$.
    In \eqref{delta},  $J(q,\hat{a})$ is nonsingular and upper bounded
    \begin{equation*}
    \|J(q,\hat{a})\| \leq \lambda_{\hat{J}}
    \end{equation*}
    for all admissible $q$.


\end{itemize}

\begin{remark}
Condition {\bf P4}  describes quantitatively the accuracy of the system parameters, whose value will affect the choice of design parameters in the controller.
The matrix $J(q,\hat{a})$ is also known as the  approximate Jacobian matrix in manipulator task-space control literature  \cite{Cheah2003approximate}.
\end{remark}\vspace{0.2cm}

By adding \eqref{im} to \eqref{law-I}, we consider the following controller 
\begin{equation}\label{law-II}
\begin{cases}
u_{i} = - K_{P}J_{i}^{\t}(q_{i},\hat{a}_{i}) \widehat{e}_{i} - K_{D}\xi_{i} + K_{I}\eta_{i} \\
\dot{\eta}_{i} = -K_{I}\eta_{i} + u_{i}
\end{cases}
\end{equation}
where $\hat{a}$ is the nominal (approximate) value of actual parameters $a$, and $K_{P}$, $K_{D}$, $K_{I}$ are positive gains to be designed.
By substituting \eqref{law-II} to 
\eqref{sys-dyna}, and using coordinate transformations and \eqref{ek}, the error dynamics of $(e,\xi)$ satisfies
\begin{equation}\label{cls-II}
\left\{
\begin{aligned}
\dot{e} &= D_{z}^{T} \overline{B}^{T}J(q,a)\xi \\ 
\dot{\xi} &= H^{-1}(q,w)[-K_{P}J^{T}(q,\hat{a})\widehat{e} - K_{D}\xi + K_{I}\eta   \\
&\quad  - C(q,\xi,w)\xi - G(q,w)]
\end{aligned}
\right.
\end{equation}
where $J(q,\hat{a})$ is the nominal (approximate) Jacobian matrix and
$\dot{\hat{a}} = \mathbf{0}$.

Now, the closed-loop error system is composed of \eqref{im-e} and \eqref{cls-II}.
Using $U_{1}(t,e,\xi)$ as in \eqref{U1-1}, it can be computed that 
\begin{align*}
\dot{U}_{1}|_{\eqref{cls-II}} &= e^{T} (K_{P} + \alpha K_{D}) D_{z}^{T} \overline{B}^{T}J(q, a)\xi \nnum\\
&\quad + \xi^{T}[-K_{P}J^{T}(q,\hat{a})\widehat{e} -K_{D}\xi+ K_{I}\eta - G(q,w)]  \nnum\\
&\quad + \alpha \dot{\widehat{e}}J(q,a)H(q,w)\xi + \alpha \widehat{e}^{T}\dot{J}(q,a)H(q,w)\xi \nnum\\
&\quad  + \alpha \widehat{e}J(q,a)\dot{H}(q,w)\xi + \alpha \widehat{e}^{T}J(q,a)[-K_{P}J^{T}(q,\hat{a})\widehat{e}  \nnum\\
&\quad   - K_{D}\xi - C(q,\xi,w)\xi  + K_{I}\eta - G(q,w)].
\end{align*}
By removing common terms in the above equation, we have
\begin{align}\label{U2-U1a}
\dot{U}_{1}|_{\eqref{cls-II}}
&= - \alpha\widehat{e}^{T}J(q(t),\hat{a})K_{P}J^{T}(q(t),\hat{a})\widehat{e} - \xi^{T}K_{D}\xi  \nnum\\
&\quad + \alpha\widehat{e}^{T}J(q(t),\tilde{a})K_{P}J^{T}(q(t),\hat{a})\widehat{e} -  K_{P}\widehat{e}^{T}J(q(t),\tilde{a})\xi  \nnum\\
&\quad + \phi_{21}(t,\tilde{\eta},e,\xi,w) + \alpha \phi_{22}(t,\tilde{\eta},e,\xi,w)
\end{align}
where
\begin{align}\label{phi-22}
\phi_{21}(t,\tilde{\eta},e,\xi,w) &= \xi^{T}\Big[ K_{I}(\tilde{\eta}+ \eta^{*} + H(q(t),w)\xi) \nnum \\
& \qquad - G(q(t),w)\Big] \nnum\\
\phi_{22}(t,\tilde{\eta},e,\xi,w) 
&= \phi_{1}(t,e,\xi,w) + \widehat{e}^{T}J(q(t),a)\Big[K_{I}(\tilde{\eta}  \nnum\\
&\quad  + \eta^{*} + H(q(t),w)\xi) - G(q(t),w)\Big]
\end{align}
with $\phi_{1}(t,e,\xi,w)$ be as in \eqref{phi-1}.
For the time derivative \eqref{U2-U1a}, we can state the following:
\begin{itemize}
    \item Using $\widehat{e} = 2\overline{B}D_{z}e$ and the Young's inequality, the second line of \eqref{U2-U1a} satisfies
    \begin{align*}
    &\alpha\widehat{e}^{T}J(q(t),\tilde{a})K_{P}J^{T}(q(t),\hat{a})\widehat{e} -  K_{P}\widehat{e}^{T}J(q(t),\tilde{a})\xi \\
    &\leq 4\alpha K_{P} \underbrace{\|J(q(t),\hat{a})\|}_{\leq \lambda_{\hat{J}}} \underbrace{\|J(q(t),\tilde{a})\|}_{\leq \delta} \|\overline{B}D_{z}e\|^{2} \\
    &\quad + 2K_{P} \underbrace{\|J(q(t),\tilde{a})\|}_{\leq \delta} \|\overline{B}D_{z}e\| \|\xi\| \\
    &\leq 4\alpha K_{P} \lambda_{\hat{J}} \delta \lambda_{3}\|e\|^{2} + K_{P}\delta(\lambda_{3}\|e\|^{2} + \|\xi\|^{2})
    \end{align*}
    where $\lambda_{3} := \lambda_{\max}\{D_{z}\overline{B^{T}B}D_{z}\}$, for all $e\in\mathcal{Q}$.

    \item It can be shown that $\phi_{21}(t,\tilde{\eta},e,\xi,w)$ and $\phi_{22}(t,\tilde{\eta},e,\xi,w)$ are smooth functions satisfying $\phi_{22}(t,0,0,0,w) = 0$, $\phi_{22}(t,0,0,0,w) = 0$ for all $t$ 
    and $w\in\mathbb{W}$. Then, 
    similar to the one given in Appendix, there are positive constants $k_{ij}$, $i=3,4, j=1,2,3$ such that
    \begin{align}
    \|\phi_{21}(t,\tilde{\eta},e,\xi,w)\| &\leq k_{31}\|\tilde{\eta}\|^{2} + k_{32}\|e\|^{2}  + k_{33}\|\xi\|^{2}  \nnum\\
    \|\phi_{22}(t,\tilde{\eta},e,\xi,w)\| &\leq k_{41}\|\tilde{\eta}\|^{2} + k_{42}\|e\|^{2}  + k_{43}\|\xi\|^{2} \label{phi22-Young}
    \end{align}
    for all for all $t$, 
    all $e\in\mathcal{Q}$, $\xi\in\Xi$ and all $w\in\mathbb{W}$.
\end{itemize}


\begin{proposition}\label{prop}
Consider the closed-loop system given by 
\eqref{im-e} and \eqref{cls-II} with sufficiently small $\delta$ in \eqref{delta}. Then for any given $K_I>0$ there exist $K_{P,\text{min}}>0$ and $\beta_{\text{min}}$ 
such that \eqref{im-e} and \eqref{cls-II} is locally exponentially stable at $(\tilde{\eta},e,\xi)=(\mathbf{0},\mathbf{0},\mathbf{0})$ with $K_P>K_{P,\text{min}}$ and $K_D/K_{P}>\beta_{\text{min}}$.
\end{proposition}\vspace{0.2cm}

\textit{Proof of Proposition~\ref{prop}:}
Define a time-varying Lyapunov function candidate $U_{2}(t,\tilde{\eta},e,\xi)$ by
\begin{equation}\label{U2}
U_{2}(t,\tilde{\eta},e,\xi) = \varepsilon^{-1} V_{\eta}(\tilde{\eta}) + U_{1}(t,e,\xi)
\end{equation}
where $V_{\eta}(\tilde{\eta})$ and $U_{1}(t,e,\xi)$ are defined in \eqref{V-eta} and \eqref{U1-1}, respectively, for a sufficiently small constant $\varepsilon>0$ to be determined later below (c.f. \eqref{varepsilon}). It is obvious that $U_{2}$ is locally positive definite if $U_{1}$ is locally positive definite as in the proof of Theorem~\ref{thm-I}. Its time derivative along the trajectory of \eqref{im-e} and \eqref{cls-II} satisfies
\begin{align*}
\dot{U}_{2}|_{\eqref{im-e}+\eqref{cls-II}}
&\leq - \Big(\frac{1}{2}\varepsilon^{-1} - k_{31} - \alpha k_{41} \Big)\|\tilde{\eta}\|^{2} - \Big(\alpha \lambda_{4} K_{P} \\
&\quad - \varepsilon^{-1} k_{21}  - k_{32} - \alpha k_{42} - 4\alpha K_{P} \lambda_{\hat{J}}\lambda_{3}\delta\Big)\|e\|^{2} \\
&\quad - (K_{D} - \varepsilon^{-1} k_{22} - k_{33} - \alpha k_{43} - K_{P}\lambda_{3}\delta)\|\xi\|^{2}
\end{align*}
for all $e\in\mathcal{Q}$ and all $\hat{w}$, where
\begin{equation}\label{lambda4}
\lambda_4 = \min_{q(t)\in \mathbb{Q}_{\epsilon}}\lambda_{\min}\{D_{z}^{T}\overline{B}^{T}J(q(t),\hat{a})J^{T}(q(t),\hat{a})\overline{B}D_{z}\}
\end{equation}
for some $\epsilon > 0$.
The above time derivative can be made negative definite by the following steps:
\begin{enumerate}[1)]
    \item Fix the constant $\alpha$ such that $0 < \alpha < \frac{c_{\min}}{c_{\max}}$.
    \item Choose any $K_{I} > 0$.
    \item Compute $k_{21}$, $k_{22}$, $k_{23}$, $k_{24}$, $\lambda_{3}$ and $\lambda_{\hat{J}}$ for some $r_{1},r_{2}>0$.
    \item Fix the constant $\varepsilon$ such that
    \begin{equation}\label{varepsilon}
    \frac{1}{2}\varepsilon^{-1} - k_{23} - \alpha k_{24} > 1.
    \end{equation}
    \item Let $\delta^{*} = \frac{\lambda_{4}}{4\lambda_{\hat{J}}\lambda_{3}}$.
    \begin{itemize}
        \item If $\delta < \delta^{*}$, then choose $K_{P}$ such that
        \begin{equation*}
        K_{P}(\alpha \lambda_{4} - 4\alpha\lambda_{\hat{J}}\lambda_{3}\delta) - \varepsilon^{-1} k_{21}  - k_{32} - \alpha k_{42} > 1.
        \end{equation*}
        \item[] Choose $K_{D}$ such that
        \begin{equation*}
        K_{D} - \varepsilon^{-1} k_{22} - k_{33} - \alpha k_{43} - K_{P}\lambda_{3}\delta > 1.
        \end{equation*}

        \item If $\delta \geq \delta^{*}$, then we can not find $K_{P}$ and $K_{D}$ such that the time derivative is negative.
    \end{itemize}
\end{enumerate}
The rest proof is similar to part~3 of the proof of Theorem~\ref{thm-I}.
Therefore, the closed-loop system \eqref{im-e} and \eqref{cls-II} is locally exponentially stability at $(\tilde{\eta},e,\xi) = (\mathbf{0},\mathbf{0},\mathbf{0})$. 

\hfill Q.E.D.

\medskip

\subsubsection{Adaptive Jacobian approach}

As mentioned before, the control law \eqref{law-II} requires that the mismatch between actual Jacobian and nominal Jacobian is bounded and sufficiently small for guaranteeing the asymptotic stability. In order to relax this, we present an adaptive Jacobian approach 
in the following theorem.

\begin{theorem}\label{thm-III}
Consider $N$ robot manipulators \eqref{sys-dyna} satisfying assumptions \textbf{P1}, \textbf{P2} and \textbf{P3}. Assume that the formation graph is  infinitesimally and minimally rigid graph $\G$.  
Then for any end-effector reference configuration $x^*$, the end-effector formation control problem can be solved by the following distributed control law, for $i=1,\ldots,N$,
\begin{equation}\label{law-III}
\begin{cases}
u_{i} = - K_{P}J_{i}^{\t}(q_{i},\hat{a}_{i}) \widehat{e}_{i} - K_{D}\xi_{i} + K_{I}\eta_{i} \\
\dot{\hat{a}}_{i} = -Z_{i}^{T}(q_{i},\widehat{e}_{i}) \big[\alpha Z_{i}(q_{i},\widehat{e}_{i})\hat{a}_{i} - \xi_{i} \big] \\
\dot{\eta}_{i} = -K_{I}\eta_{i} + u_{i}
\end{cases}
\end{equation}
with design parameters $\alpha$, $K_{P}$, $K_{D}$ and $K_{I}$, where $\widehat{e}_{i}$, $\xi_{i}$ are given in \eqref{ei}, \eqref{xi}, respectively.
In particular, there are constants $\alpha,K_{P,\text{min}},K_{D,\text{min}},K_{I},r,\epsilon, \varepsilon>0$ for the closed-loop system \eqref{sys-dyna}, \eqref{law-III} such that if $K_{P}\geq K_{P,\text{min}}$, $K_{D}\geq K_{D,\text{min}}$, $(\tilde{a}(0),\tilde{\eta}(0), e(0),\xi(0))\in  \{\tilde{a},\tilde{\eta},e,\xi: \|\tilde{a}\|^{2} + \frac{\varepsilon^{-1}K_{I}^{-1}}{2}\|\tilde{\eta}\|^{2} + \frac{K_{P}+2\alpha K_{D}}{4}\|e\|^{2} + \frac{c_{\min}+c_{\max}}{4}\|\xi\|^{2} \leq r\}$ and $q(0)\in\mathbb{Q}_{\epsilon} := \{q : \text{dist} (q,\mathrm{Q}) > \epsilon \}$ then the closed-loop system is stable and state $(\tilde{\eta},e,\xi)$ converges to zero asymptotically.
\end{theorem}\vspace{0.2cm}

\textit{Proof of Theorem~\ref{thm-III}:}
Substituting \eqref{law-III} into \eqref{sys-dyna}, the closed-loop system is composed of subsystem
$(e,\xi)$ as in \eqref{cls-II}, subsystem $\tilde{\eta}$ as in \eqref{im-e}, and the following system describing parameter estimation error
\begin{equation}\label{e-est}
\dot{\tilde{a}} = -Z^{T}(q,\widehat{e}) \big[\alpha Z(q,\widehat{e})\hat{a} - \xi \big]
\end{equation}
where $\tilde{a} = \hat{a} - a$, $\hat{a}=\col(\hat{a}_{1},\ldots,\hat{a}_{N})$, $a=\col(a_{1},\ldots,a_{N})$, and $Z(q,\widehat{e}) = \text{block diag}(Z(q_{1},\widehat{e}_{1}),\ldots,Z(q_{N},\widehat{e}_{N}))$.

Next, before carrying out the Lyapunov analysis, we show the following parameter linearization conditions for the second line of \eqref{U2-U1a}.
\begin{itemize}
    \item Since the manipulator Jacobian matrix satisfies linear parameterized condition {\bf P3}, $\alpha\widehat{e}^{T}J(q,\tilde{a})K_{P}J^{T}(q,\hat{a})\widehat{e}$ can be rewritten as
    \begin{align}\label{L1}
    \alpha\widehat{e}^{T}J(q,\tilde{a})K_{P}J^{T}(q,\hat{a})\widehat{e}
    &= \alpha K_{P}[J^{T}(q,\tilde{a})\widehat{e}]^{T} J^{T}(q,\hat{a})\widehat{e} \nnum\\
    &= \alpha K_{P}[Z(q,\widehat{e})\tilde{a}]^{T} Z(q,\widehat{e})\hat{a} \nnum\\
    &= \alpha K_{P}\tilde{a}^{T}Z^{T}(q,\widehat{e}) Z(q,\widehat{e})\hat{a}.
    \end{align}

    \item Using {\bf P3} again, $-K_{P}\widehat{e}^{T}J(q,\tilde{a})\xi$ can be rewritten as
    \begin{align}\label{L2}
    -  K_{P}\widehat{e}^{T}J(q,\tilde{a})\xi &= -  K_{P}[J^{T}(q,\tilde{a})\widehat{e}]^{T}\xi \nnum\\
    &= - K_{P}[Z(q,\widehat{e})\tilde{a}]^{T}\xi \nnum\\
    &= - K_{P}\tilde{a}^{T}Z^{T}(q,\widehat{e}) \xi.
    \end{align}
\end{itemize}

Let $U_{3}:=U_{3}(t,\tilde{a},\tilde{\eta},e,\xi)$ be a Lyapunov function candidate defined by
\begin{align}\label{U3-d-1}
U_{3} = \frac{1}{2}\tilde{a}^{T}K_{P}\tilde{a} + U_{2}(t,\tilde{\eta},e,\xi)
\end{align}
where $U_{2}(t,\tilde{\eta},e,\xi)$ is defined in \eqref{U2}.
Then the time derivative of $U_{3}$ along the trajectory of the closed-loop system composed of \eqref{im-e}, \eqref{cls-II} and \eqref{e-est} satisfies
\begin{align*}
&\dot{U}_{3}|_{\eqref{im-e}+\eqref{cls-II} +\eqref{e-est} } \\
&= \dot{U}_{2}|_{\eqref{im-e}+\eqref{cls-II}} - K_{P}\tilde{a}^{T}Z^{T}(q(t),\widehat{e}) \big[\lambda Z(q(t),\widehat{e})\hat{a} - \xi \big] \\
&= \dot{V}_{\eta}|_{\eqref{im-e}} + \dot{U}_{1}|_{\eqref{cls-II}} - K_{P}\tilde{a}^{T}Z^{T}(q(t),\widehat{e}) \big[\lambda Z(q(t),\widehat{e})\hat{a} - \xi \big]
\end{align*}
Then by using \eqref{V-eta-d} and \eqref{U2-U1a}, we obtain
\begin{align}
&\dot{U}_{3}|_{\eqref{im-e}+\eqref{cls-II} +\eqref{e-est} } \nnum\\
&= - \frac{1}{2}\varepsilon^{-1}\tilde{\eta}^{T}\tilde{\eta} - \alpha\widehat{e}^{T}J(q(t),\hat{a})K_{P}J^{T}(q(t),\hat{a})\widehat{e} - \xi^{T}K_{D}\xi  \nnum\\
&\quad + \phi_{21}(t,\tilde{\eta},e,\xi,w) + \alpha \phi_{22}(t,\tilde{\eta},e,\xi,w)
\end{align}
where functions $\phi_{22}$ and $\phi_{22}$ are the same as those in \eqref{U2-U1a}. Using the growth condition of $\phi_{22}$ and $\phi_{22}$, we have
\begin{align*}
\dot{U}_{3}|_{\eqref{im-e}+\eqref{cls-II} +\eqref{e-est} } &\leq - \Big(\frac{1}{2}\varepsilon^{-1} - k_{31} - \alpha k_{41} \Big)\|\tilde{\eta}\|^{2}  \\
&\quad - (\alpha \lambda_{4} K_{P} - \varepsilon^{-1} k_{21}  - k_{32} - \alpha k_{42})\|e\|^{2} \\
&\quad - (K_{D} - \varepsilon^{-1} k_{22} - k_{33} - \alpha k_{43} )\|\xi\|^{2}.
\end{align*}
Similar to the analysis in the proof of Proposition~\ref{prop}, we can first fix parameter $\varepsilon$ such that $\frac{1}{2}\varepsilon^{-1} - k_{23} - \alpha k_{24} > 0$. Subsequently, we can  choose $K_{D}$ and $K_{D}$ such that
\begin{equation}\label{U3-d-c}
\begin{aligned}
    \alpha \lambda_{4} K_{P} - \varepsilon^{-1} k_{21}  - k_{32} - \alpha k_{42}  > 1 \\
    K_{D} - \varepsilon^{-1} k_{22} - k_{33} - \alpha k_{43} > 1.
\end{aligned}
\end{equation}
Hence, we have
\begin{equation*}
\dot{U}_{3}|_{\eqref{im-e}+\eqref{cls-II} +\eqref{e-est}} \leq - \|\tilde{\eta}\|^{2} - \|e\|^{2} - \|\xi\|^{2}.
\end{equation*}
which implies that $U_{3}$ is non-increasing.
Since $U_{3}$ is locally positive definite, the states $(\eta,e,\xi,\tilde{a})$ of system \eqref{im-e}, \eqref{cls-II} and \eqref{e-est} starting from a small neighborhood of origin are all bounded over time interval $[0,\infty)$. Hence, by the continuity, $\ddot{U}_{3}$ along the trajectory of closed-loop system is also bounded.
Using Barbalat’s Lemma \cite[pp. 123]{Slotine-book}, it implies $\tilde{\eta}$, $e$ and $\xi$ converges to zero. Similarly as in Theorem 3.1, we can conclude that $x(t)\to\mathcal{S}_W$ as $t\to\infty$ with manipulators operating away from kinematic singularities.
The proof is complete.
\hfill Q.E.D. \vspace{0.2cm}


\begin{remark}
Let $p(t)$ be the geometric centroid of the formation by $p(t) = \frac{1}{N}\sum_{i=1}^{N}x_{i}(t)$.
Unlike the point mass model case \cite{Oh2014distance}, this centroid is not necessarily stationary under the proposed control law even when the manipulator parameters are all perfectly known.
\end{remark}

\begin{remark}
The internal model $\eta_{i}$ of \eqref{law-III} is essentially an integrator \cite[Chapter~12.3]{Khalil-book}. This property can be shown by a direct coordinate transformation. By substituting $u_{i}$ of \eqref{law-III} into subsystem $\eta_{i}$, direct calculation gives
\begin{align*}
\dot{\eta}_{i} &= -K_{I}\eta_{i} + [- K_{P}J_{i}^{\t}(q_{i},\hat{a}_{i}) \widehat{e}_{i} - K_{D}\xi_{i} + K_{I}\eta_{i}] \\
&= -K_{P}J_{i}^{\t}(q_{i},\hat{a}_{i}) \widehat{e}_{i} - K_{D}\xi_{i}.
\end{align*}
Define an output vector $y_{i}$ by
\begin{equation*}
y=\Lambda J_{i}^{\t}(q_{i},\hat{a}_{i}) \widehat{e}_{i} + \xi_{i}
\end{equation*}
where $\Lambda = K_{D}^{-1}K_{P}$.
Then the input $u_{i}$ of \eqref{law-III} can be rewritten as
\begin{equation*}
u_{i} = -K_{P}J_{i}^{\t}(q_{i},\hat{a}_{i}) \widehat{e}_{i} - K_{D}\xi_{i} - \bar{K}_{I}\int_{0}^{t}y(s)ds 
\end{equation*}
where $\bar{K}_{I}=K_{I}K_{D}$, {\sl cf.} the PID form in \cite{Cheah1999pid,Cheah2003approximate}.
\end{remark}

\begin{figure}
\centering
\begin{tikzpicture}[scale=0.45]

\begin{scope}[xshift = 0cm, yshift=0cm] 
\robotbasenewS 
\node[mark size=2pt,color=red] (base1) {\pgfuseplotmark{*}};
\draw [] (0,-1) node{\color{black}\footnotesize Agent 1};
\link(\thetaoneA:\LoneA);
\joint
\begin{scope}[shift=(\thetaoneA:\LoneA), rotate=\thetaoneA]
    \link(\thetatwoA:\LtwoA);
    \joint
    \begin{scope}[shift=(\thetatwoA:\LtwoA), rotate=\thetatwoA]
        \grip
        \node[mark size=2pt,color=red] (agent1) {\pgfuseplotmark{*}};
    \end{scope}
\end{scope}

\begin{scope}[xshift = 10cm, yshift=1.2cm] 
    \begin{scope}[rotate=30]
    \robotbasenewS
    \node[mark size=2pt,color=red] (base2) {\pgfuseplotmark{*}};
    \end{scope}
    \draw [] (0,-1) node{\color{black}\footnotesize Agent 2};
    \link(\thetaoneB:\LoneB);
    \joint
    \begin{scope}[shift=(\thetaoneB:\LoneB), rotate=\thetaoneB]
        \link(\thetatwoB:\LtwoB);
        \joint
        \begin{scope}[shift=(\thetatwoB:\LtwoB), rotate=\thetatwoB]
            \grip
            \node[mark size=2pt,color=red] (agent2) {\pgfuseplotmark{*}};
       \end{scope}
    \end{scope}
\end{scope}

\begin{scope}[xshift=8.8cm, yshift=8.66cm] 
    \begin{scope}[rotate=60]
    \robotbasenewS
    \node[mark size=2pt,color=red] (base3) {\pgfuseplotmark{*}};
    \end{scope}
    \draw [] (0,-1) node{\color{black}\footnotesize Agent 3};
    \link(\thetaoneD:\LoneD);
    \joint
    \begin{scope}[shift=(\thetaoneD:\LoneD), rotate=\thetaoneD]
        \link(\thetatwoD:\LtwoD);
        \joint
        \begin{scope}[shift=(\thetatwoD:\LtwoD), rotate=\thetatwoD]
            \grip
            \node[mark size=2pt,color=red] (agent3) {\pgfuseplotmark{*}};
        \end{scope}
    \end{scope}
\end{scope}

\begin{scope}[xshift=-1.2cm, yshift=7.46cm] 
    \begin{scope}[rotate=-30]
    \robotbasenewS
    \node[mark size=2pt,color=red] (base4) {\pgfuseplotmark{*}};
    \end{scope}
    \draw [] (0,-1) node{\color{black}\footnotesize Agent 4};
    \link(\thetaoneC:\LoneC);
    \joint
    \begin{scope}[shift=(\thetaoneC:\LoneC), rotate=\thetaoneC]
        \link(\thetatwoC:\LtwoC);
        \joint
        \begin{scope}[shift=(\thetatwoC:\LtwoC), rotate=\thetatwoC]
            \grip
            \node[mark size=2pt,color=red] (agent4) {\pgfuseplotmark{*}};
        \end{scope}
    \end{scope}
\end{scope}

\draw [spring,gray!40] (agent1) -- node[below] {\color{gray}$z_{1}$} (agent2);
\draw [spring,gray!40] (agent2) -- node[above right] {\color{gray}$z_{2}$} (agent3);
\draw [spring,gray!40] (agent3) -- node[above left] {\color{gray}$z_{3}$} (agent4);
\draw [spring,gray!40] (agent4) -- node[above left] {\color{gray}$z_{4}$} (agent1);
\draw [spring,gray!40] (agent1) -- node[above left] {\color{gray}$z_{5}$} (agent3);

\draw[-latex,thick,red,rotate=0] ($(base1)+(0,0)$) -- ($(base1)+(1.8,0)$);
\draw[-latex,thick,red,rotate=0] ($(base1)+(0,0)$) node[right, yshift=0.5cm] {\footnotesize\color{red}$\Sigma_{1}$} -- ($(base1)+(0,1.8)$);
\draw[-latex,thick,red,rotate=30] ($(base2)+(0,0)$) -- ($(base2)+(1.8,0)$);
\draw[-latex,thick,red,rotate=30] ($(base2)+(0,0)$) node[yshift=0.5cm] {\footnotesize\color{red}$\Sigma_{2}$} -- ($(base2)+(0,1.8)$);
\draw[-latex,thick,red,rotate=60] ($(base3)+(0,0)$) -- ($(base3)+(1.8,0)$);
\draw[-latex,thick,red,rotate=60] ($(base3)+(0,0)$) node[yshift=0.5cm] {\footnotesize\color{red}$\Sigma_{3}$} -- ($(base3)+(0,1.8)$);
\draw[-latex,thick,red,rotate=-30] ($(base4)+(0,0)$) -- ($(base4)+(1.8,0)$);
\draw[-latex,thick,red,rotate=-30] ($(base4)+(0,0)$) node[yshift=0.5cm] {\footnotesize\color{red}$\Sigma_{4}$} -- ($(base4)+(0,1.8)$);

\begin{scope}[xshift=-3cm, yshift=3cm] 
    \node[mark size=2pt,color=red] (world) {\pgfuseplotmark{*}};
\draw[-latex,thick,red] ($(world)+(0,0)$) node[left] {\footnotesize\color{red}$\Sigma_{g}$} -- ($(world)+(1.8,0)$) node[below] {\footnotesize\color{red}$X$};
\draw[-latex,thick,red] ($(world)+(0,0)$) -- ($(world)+(0,1.8)$) node[left] {\footnotesize\color{red}$Y$} ;
\end{scope}


\end{scope}

\end{tikzpicture}
\caption{Distanced-based end-effector formation of 4 two-link planar manipulators. $\Sigma_{g}$ is the global frame. $\Sigma_{1}$ to $\Sigma_{4}$ are local frames fixed to the base of each manipulator. These local coordinate systems do not need to have a common sense of orientation,}\label{fig-dist-robots}
\end{figure}
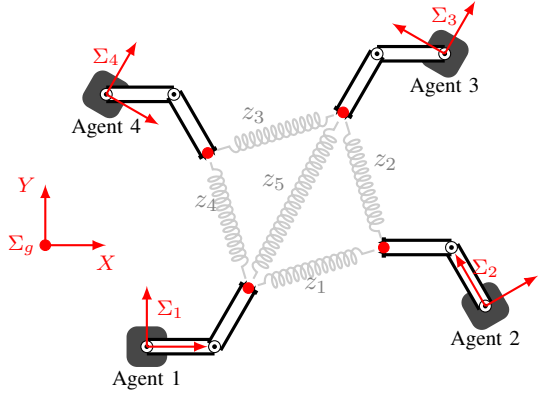

\section{Further Discussion}\label{sec-disc}

The proposed gradient-based designs in aforementioned sections are applicable to both displacement- and distance-based formation control.
For the displacement-based formation control, the proof can be carefully modified by replacing $2\overline{B}D_{z}$ with $\overline{B}$ in the closed-loop system and relevant Lyapunov analysis due to \eqref{eq: disp} and \eqref{eq: dist}.

For the distance-based formation control, just like the gradient-based design for the kinematic point case \cite{Oh2014distance}, the manipulators can maintain their own coordinate system without the use of common frame of reference. 
In other words, the proposed distributed gradient control law can be implemented using its local frame of reference and using only local relative measurement systems, both of which are desirable in practice.

The realization using only local frame of reference can be shown by a suitable coordinate transformation.
As depicted in Fig.~\ref{fig-dist-robots}, let $\Sigma_{i}$ denote the local frame fixed to the base of the $i$th manipulator.
By adopting a group of new notation in which superscripts are used to denote local coordinate system, the manipulator dynamics and kinematics can be written as
\begin{gather*}
H_{i}(q_{i},w_{i})\ddot{q}_{i} + C_{i}(q_{i},\dot{q}_{i},w_{i})\dot{q}_{i} + G_{i}^{i}(q_{i},w_{i}) = u_{i}^{i} \\
x_{i}^{i} = h_{i}^{i}(q_{i},w_{i})
\end{gather*}
where $i\in\{1,\ldots,N\}$, $x_{i}^{i}$, $u_{i}^{i}$, $h_{i}^{i}(q_{i},w_{i})$ and $G_{i}^{i}(q_{i},w_{i})$ are the end-effector position, the control input, the forward kinematics and the gravitational torque, respectively, that defined with respect to $\Sigma_{i}$. The
states $q_{i},\dot{q}_{i},\ddot{q}_{i}\in\R^{n}$ and functions $H_{i}(q_{i},w_{i})$, $C_{i}(q_{i},\dot{q}_{i},w_{i})$ are the same as that in previous sections, because joint angle, the kinetic energy and the forward kinematic with respect to base frame are not defined with respect to the world frame $\Sigma_{g}$.

Let $x_{i}^{g}$ and $x_{i0}$ be the end-effector position and base position of $i$-th manipulator, respectively, with respect to the global frame $\Sigma_{g}$. Then we have
\begin{equation}\label{x-LG}
x_{i}^{g} = R_{i}^{g}x_{i}^{i} + x_{i0}
\end{equation}
where $R_{i}^{g}\in\text{SO($m$)}$ is a rotation matrix defining the rotation transformation from $\Sigma_{i}$ to $\Sigma_{g}$.
Taking the time derivative of \eqref{x-LG} gives us
\begin{align}\label{Ji-local}
J_{i}^{g}(q_{i},a_{i})\dot{q}_{i} &= R_{i}^{g}J_{i}^{i}(q_{i},a_{i})\dot{q}_{i}.
\end{align}
where $J_{i}^{i}(q_{i},a_{i}) = \frac{\partial h_{i}^{i}(q_{i},w_{i})}{\partial q_{i}}$.
Suppose that all the manipulators sense relative end-effector positions of their neighbors with respect to their own base frame $\Sigma_{i}$
\begin{align*}
z_{ij}^{i} = z_{k}^{i} = x_{i}^{i} - x_{j}^{i},~~ j\in\N_{i}
\end{align*}
where $x_{j}^{i}$ is the $j$th manipulator's end-effector position with respect to $\Sigma_{i}$.
Thus the error signal for the edge $k$ is $e_{ij}^{i} = e_{k}^{i} = \|z_{k}^{i}\|^{2} - \|z_{k}^{*}\|^{2}$ satisfying $e_{ij}^{i} = e_{ij}$, where $e_{ij}$ is the error signal with respect to $\Sigma_{g}$.
From \eqref{ei}, it can be expressed locally as 
\begin{equation*}
\widehat{e}_{i}^{i}  = \sum_{j\in\N_{i}} b_{ik}e_{k}^{i}(x_{i}^{i} - x_{j}^{i}) = \sum_{j\in\N_{i}} e_{ij}^{i} (x_{i}^{i} - x_{j}^{i}).
\end{equation*}
satisfying
\begin{equation}\label{ei-local}
\widehat{e}_{i} = \sum_{j\in\N_{i}} e_{ij}^{i} (x_{i} - x_{j}) = \sum_{j\in\N_{i}} e_{ij}^{i} R_{i}^{g}(x_{i}^{i} - x_{j}^{i}) = R_{i}^{g}\widehat{e}_{i}^{i}.
\end{equation}

Hence, the gradient-based control law for agent $i$ can be designed as
\begin{align*}
u_{i}^{i} &= - K_{P}[\underbrace{J_{i}^{i}(q_{i},\hat{a}_{i})}_{\text{using \eqref{Ji-local}} }]^{T} \widehat{e}_{i}^{i}  - K_{D}\xi_{i} + K_{I}\eta_{i} \\
&= - K_{P}\underbrace{[(R_{i}^{g})^{-1}J_{i}^{g}(q_{i},\hat{a}_{i})]^{T}}_{\text{using $R_{i}^{g}\in$ SO($m$)}} \widehat{e}_{i}^{i}  - K_{D}\xi_{i} + K_{I}\eta_{i} \\
&= - K_{P}[J_{i}^{i}(q_{i},\hat{a}_{i})]^{T} \underbrace{R_{i}^{g} \widehat{e}_{i}^{i}}_{\text{using \eqref{ei-local}}}  - K_{D}\xi_{i} + K_{I}\eta_{i} \\
&= - K_{P}[J_{i}^{g}(q_{i},\hat{a}_{i})]^{T} \widehat{e}_{i}  - K_{D}\xi_{i} + K_{I}\eta_{i} = u_{i}
\end{align*}
where $u_{i}$ is the input specified with respect to $\Sigma_{g}$.

\section{Simulation}\label{sec-sim}

\subsection{End-effectors formation in 2D}
For the simulation setup, we first consider a network of $N=4$ two-link planar manipulator in the horizontal X-Y plane.
For the dynamic model of two-link robot manipulator as in \eqref{sys-dyna}, we refer to \cite[Example 6.2]{Slotine-book} and the corresponding nominal values of the parameters are given in Table~\ref{table-2DOF-para} for each link. 
The kinematic model of each two-link robot manipulator is given by
\begin{align*}
x_{i} = \bbm{l_{1}\cos(q_{i1}) + l_{2}\cos(q_{i1} + q_{i2}) \\ l_{1}\sin(q_{i1}) + l_{2}\sin(q_{i1} + q_{i2})} + x_{i0}
\end{align*}
and correspondingly, the manipulator Jacobian matrix is
\begin{align*}
&J_{i}(q_{i},w_{i}) \\
&= \bbm{-l_{1}\sin(q_{i1}) - l_{2}\sin(q_{i1} + q_{i2}) & -l_{2}\sin(q_{i1} + q_{i2}) \\ l_{1}\cos(q_{i1}) + l_{2}\cos(q_{i1} + q_{i2}) & l_{2}\cos(q_{i1} + q_{i2})}
\end{align*}
for $i=1,2,3,4$, where $q_{i} = \bbm{q_{i1} & q_{i2}}^{T}$. Then the kinematic singular configurations set is given by $\{q_{i1}\in\R,q_{i2}\in\R : q_{i2} = 0,\pm \pi,\pm 2\pi,\dots\}$, $i=1,2,3,4$.

We consider the formation shape of a square with side length of $0.4$ m and the associated  formation graph is represented by its incidence matrix given by
\begin{equation*}
B = \begin{bmatrix} 1 & 0 & 0 & -1 & 1 \\ -1 & 1 & 0 & 0 & 0 \\ 0 & -1 & 1 & 0 & -1 \\ 0 & 0 & -1 & 1& 0 \end{bmatrix}
\end{equation*}
and illustrated in  Fig.~\ref{fig-robots} (right).
For the numerical simulation setup, the bases of the 4 manipulators are located at $(0,0)$, $(6,0)$, $(6,6)$ and $(0,6)$, respectively, and the initial joint positions are set to $q_{1}=[0~~\pi/3]^{T}$, $q_{2}=[\pi/2~~ \pi/3]^{T}$, $q_{3}=[\pi~~\pi/3]^{T}$, $q_{4}=[3\pi/2~~\pi/3]^{T}$. All the initial joint velocities are set to zero.
The initial values of kinematic parameter estimates are determined as $\hat{a}_{i} = \bbm{2 & 2}^{T}$, $i=1,2,3,4$.

We use the result presented in Theorem~\ref{thm-III}.
Since  all manipulators are considered to operate in the horizontal plane ($G_{i}(q_{i},w_{i})\equiv 0$, $i=1,2,3,4$), the internal model subsystem $\eta_{i}$ to compensate the gravity is not needed and $K_{I}$ is set to be zero.

In order to determine the design parameters $K_{P}$, $K_{D}$ and $\alpha$, we approximate the value of the gain parameters $k_{ij}$ in \eqref{U3-d-c} numerically by selecting a group of points sufficiently dense and properly distributed in a compact set.
For example, the value of $\lambda_{4} = \lambda_{\min}\{D_{z}^{T}\overline{B}^{T}J(q,\hat{a})J^{T}(q,\hat{a})\overline{B}D_{z}\}$ defined in \eqref{lambda4} is computed in a neighborhood of the desired shape $z^{*}$.
Specifically, we take points for each element of $z=\col(z_{1},\dots,z_{5})\in\R^{10}$ for every $0.5$ such that $\|e\| < r_{1}$ with $r_{1} = 16$, take points for each element of $q$ for every $\frac{\pi}{6}$ such that $\frac{i-1}{2}\pi \leq q_{i1} \leq \frac{1}{3}\pi + \frac{i-1}{2}\pi$, $\frac{1}{6}\pi\leq q_{i2} \leq \frac{1}{2}\pi$, $(i=1,2,3,4)$,
and take points for each element of $\hat{a}$ for every $0.2$ such that $1.5 \leq \hat{a}_{ij} \leq 2.5, \, (i=1,2,3,4, j = 1,2)$. Then, by sequentially calculating the minimum eigenvalue of $D_{z}^{T}\overline{B}^{T}J(q,\hat{a})J^{T}(q,\hat{a})\overline{B}D_{z}$ for all the points, we can approximate that $\lambda_{4} = 0.5$.
Similarly, we can estimate that $c_{i,\min} = 0.16$, $c_{i,\max} = 7.8$,
$k_{11}=450$, $k_{12}=9000$ by applying this grid method. Since there is no $\eta$-subsystem, then we have $k_{21}=0$, $k_{22}=0$, $k_{31}=0$, $k_{32}=0$, $k_{33}=0$, $k_{41}=0$, $k_{42} = k_{11}$, $k_{43}=k_{12}$.
Therefore, the controller parameters can be chosen as follows: $\alpha=0.02$, $K_{P}=800$ and $K_{D}=180$.

\begin{figure}
  \centering
  \includegraphics[width=0.5\textwidth]{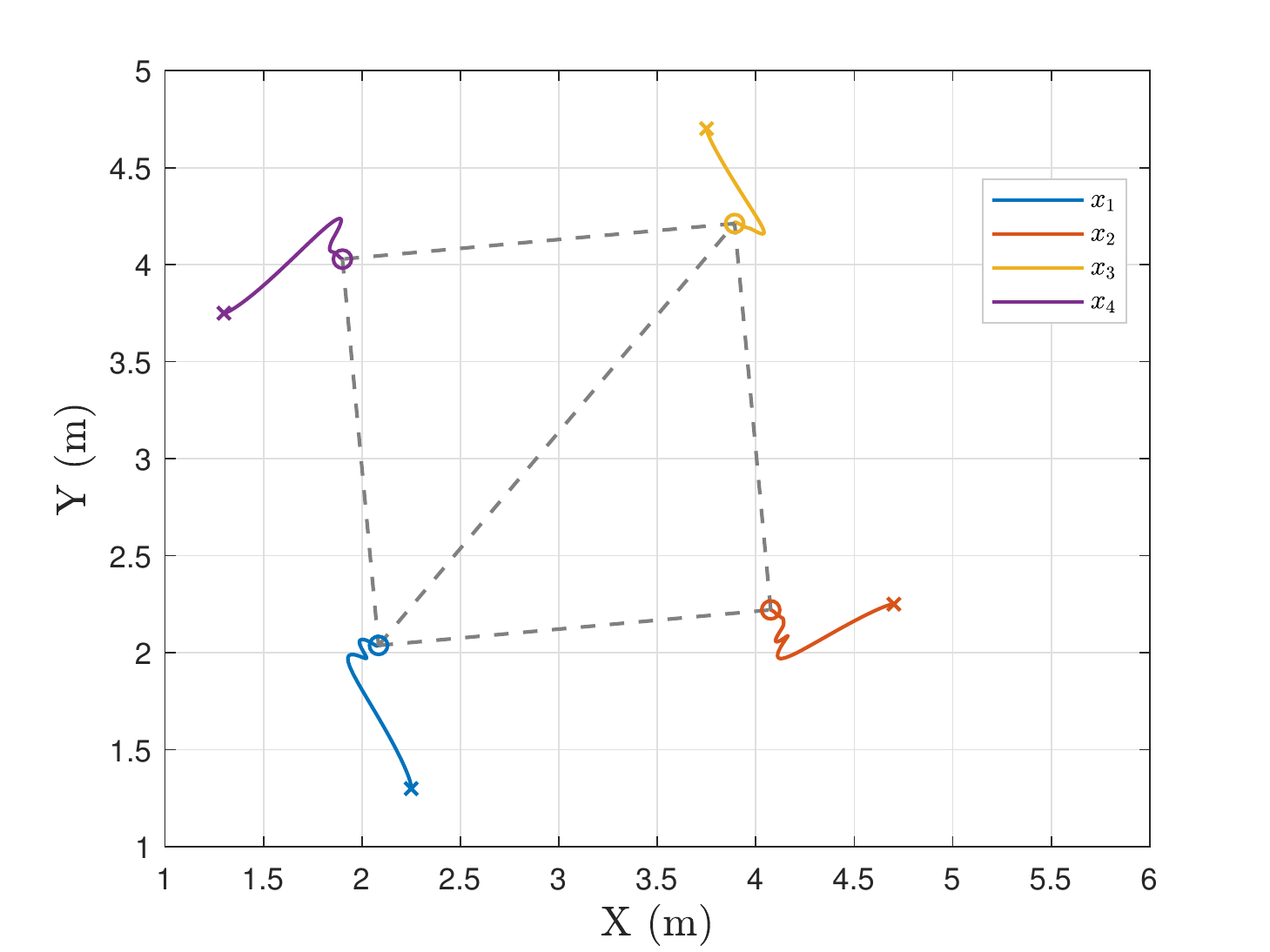}
  \caption{Trajectories of the manipulator end-effectors from the initial positions ($\times$) to the final positions ($\circ$) in 2D space.}\label{fig-x-2D}
\end{figure}

\begin{figure}
  \centering
  \includegraphics[width=0.5\textwidth]{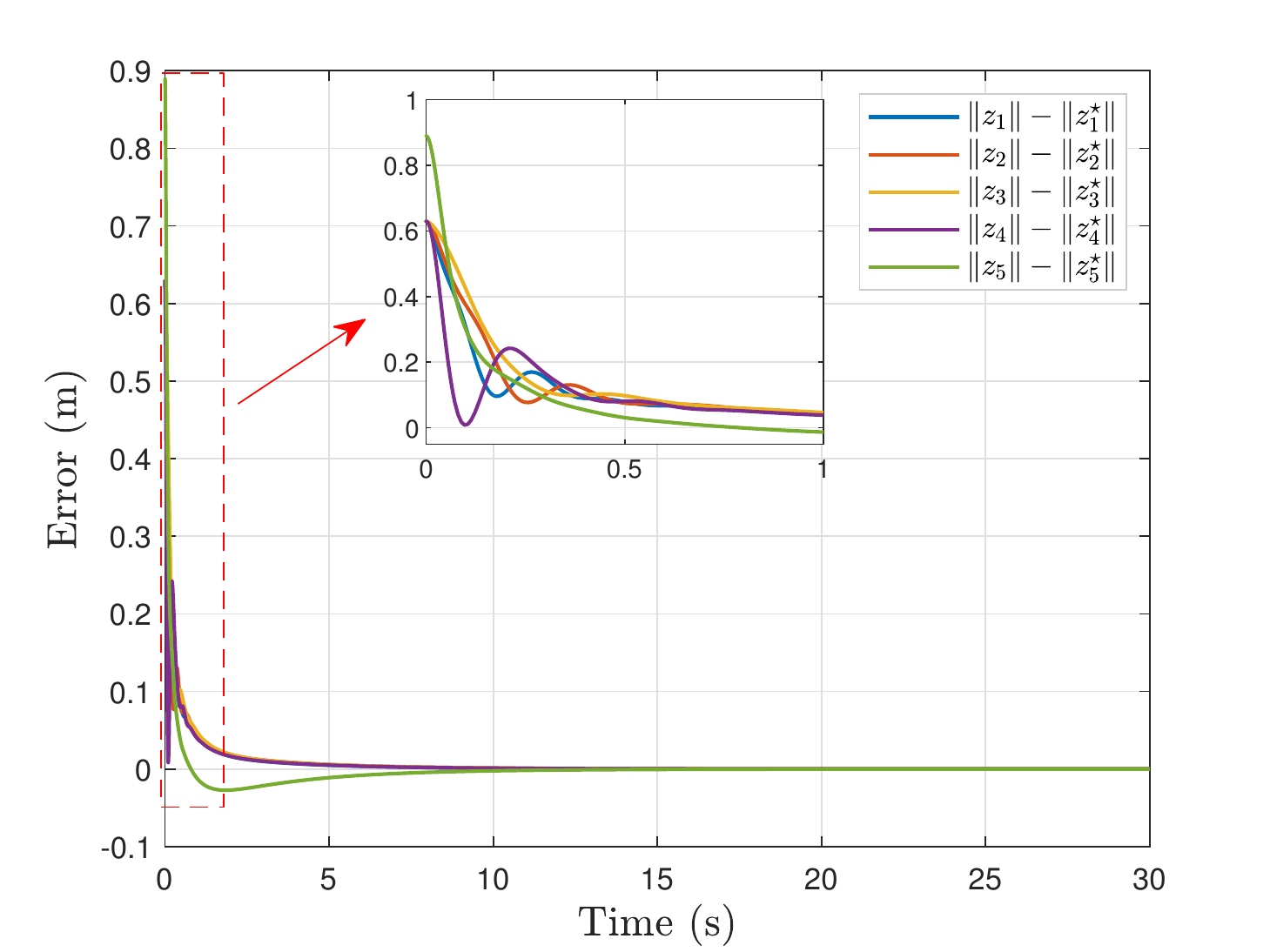}
  \caption{Performance of the inner distance error in 2D space.}\label{fig-e-2D}
\end{figure}

\begin{figure}
  \centering
  \includegraphics[width=0.5\textwidth]{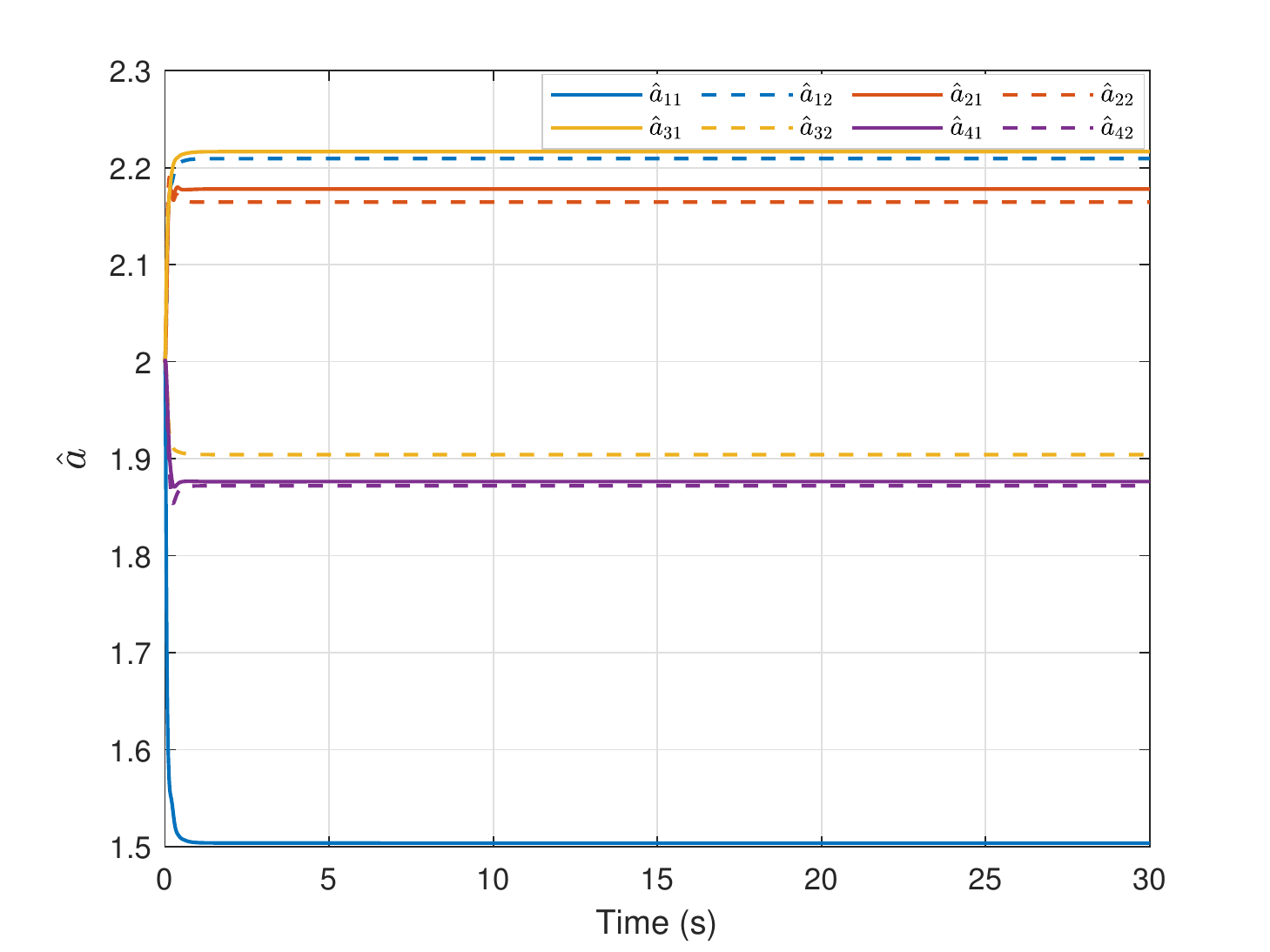}
  \caption{Kinematic parameter estimates in 2D space.}\label{fig-est-2D}
\end{figure}

\begin{figure}
  \centering
  \includegraphics[width=0.5\textwidth]{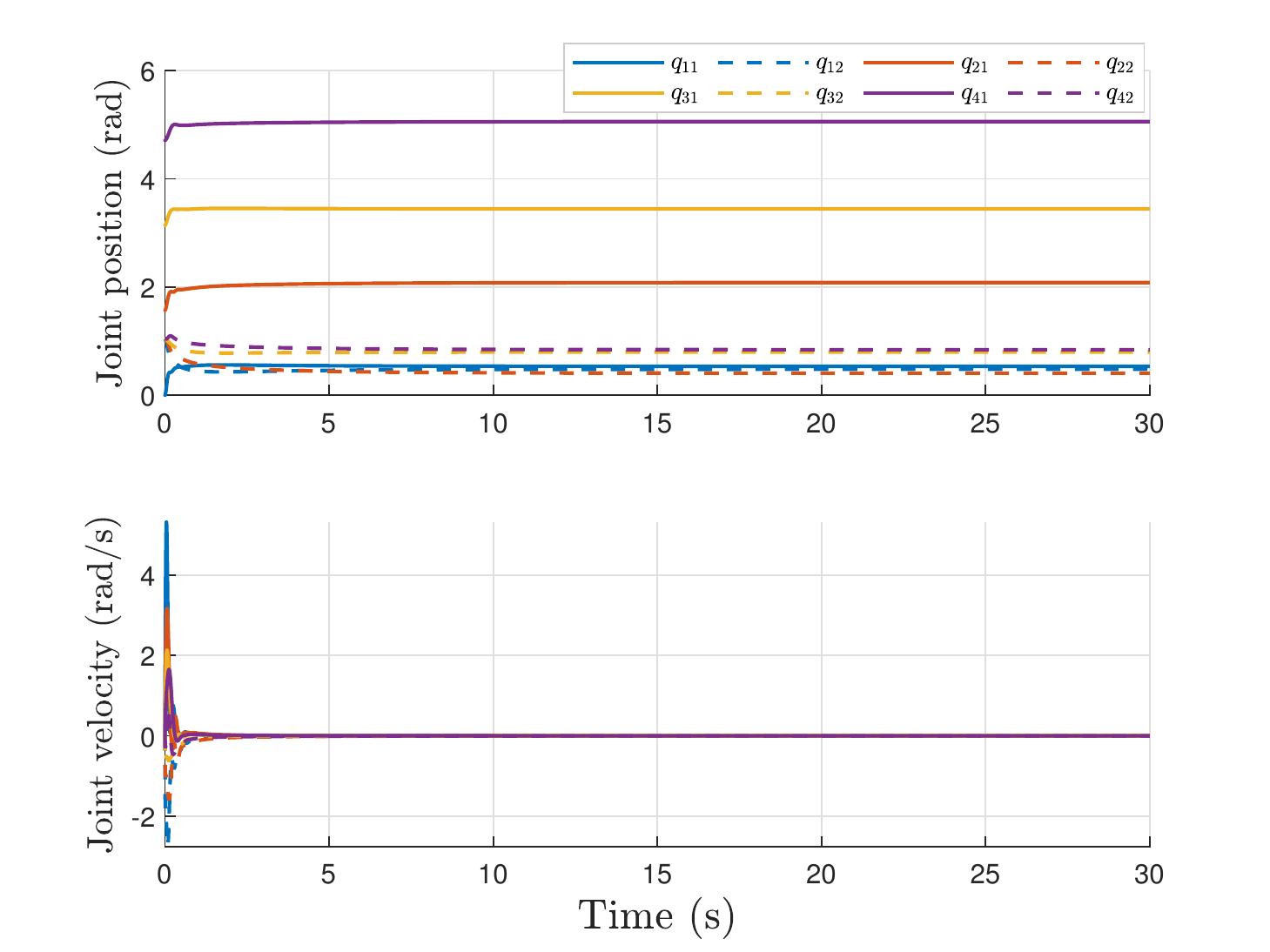}
  \caption{Performance of joint trajectories and velocities in 2D space.}\label{fig-joint-2D}
\end{figure}

We run the simulation for $30$ seconds until the formation converges and the simulation results are shown in Figures \ref{fig-x-2D} to \ref{fig-joint-2D}. The trajectories and formation pattern of the manipulators' end-effector as presented 
in Fig.~\ref{fig-x-2D}. Fig.~\ref{fig-e-2D} shows that the inner distance errors converge to zero as expected.  
Fig.~\ref{fig-est-2D} shows the evaluation of kinematic parameter estimates.
From Fig.~\ref{fig-joint-2D}, where the joint positions and velocities are plotted, we can conclude that the end-effectors remain stationary once they reach the intended shape, e.g., they do not exhibit undesirable group motion.

\begin{table}
  \caption{\small Physical parameters of the two-link planar manipulators. }\label{table-2DOF-para}
  \centering{\footnotesize
  \begin{tabular}{cccc}
    \toprule
    \multirow{2}{*}{Symbol} & \multirow{2}{*}{Meaning} & \multicolumn{2}{c}{Nominal value} \\
    \cmidrule(r){3-4}
    & & $i=1$ & $i=2$ \\
    \midrule
    $m_{i}$ (Kg) & mass of the $i$th link & $1.2$ & $1.0$ \\
    \midrule
    $I_{ci}$ (Kg$\cdot$m$^{2}$) & $i$th moment of inertia & 0.2250 & 0.1875 \\
    \midrule
    $l_{i}$ (m) & length of the $i$th link & 1.5 & 1.5 \\
    \midrule
    $l_{ci}$ (m) & \tabincell{c}{distance from the center of the \\ mass of the $i$th link to the $i$th joint} & 0.75 & 0.75 \\
    \bottomrule
  \end{tabular}}
\end{table}


\begin{figure}
  \centering
  \includegraphics[width=0.2\textwidth]{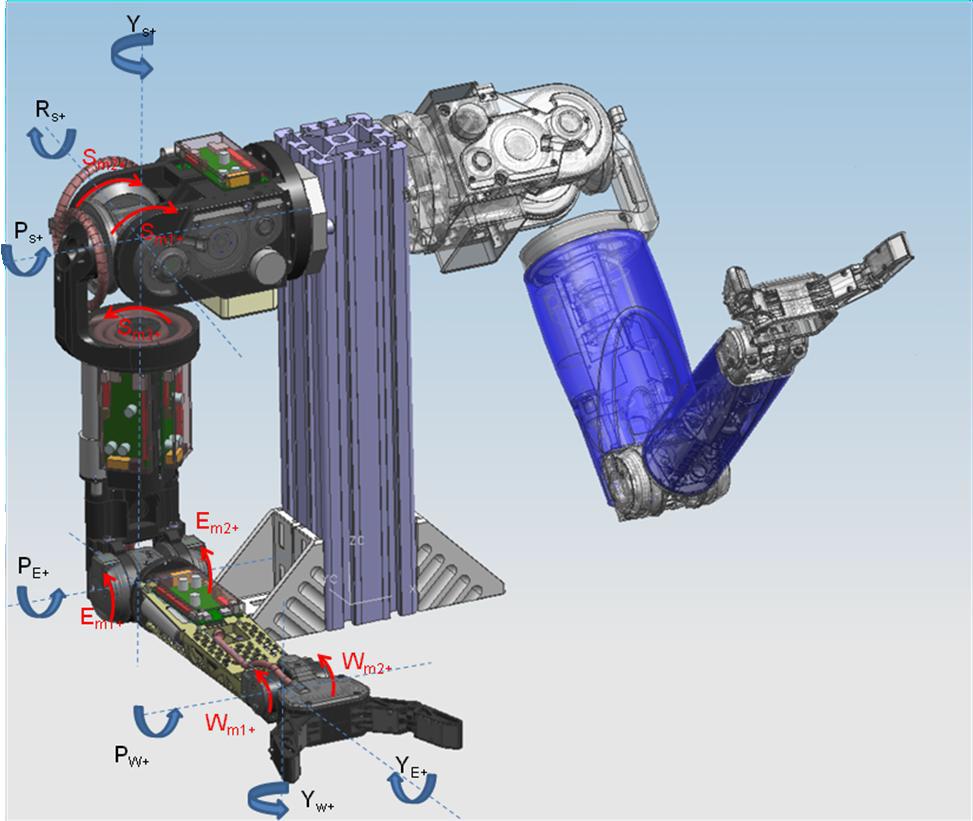}
   \,
  \includegraphics[width=0.25\textwidth]{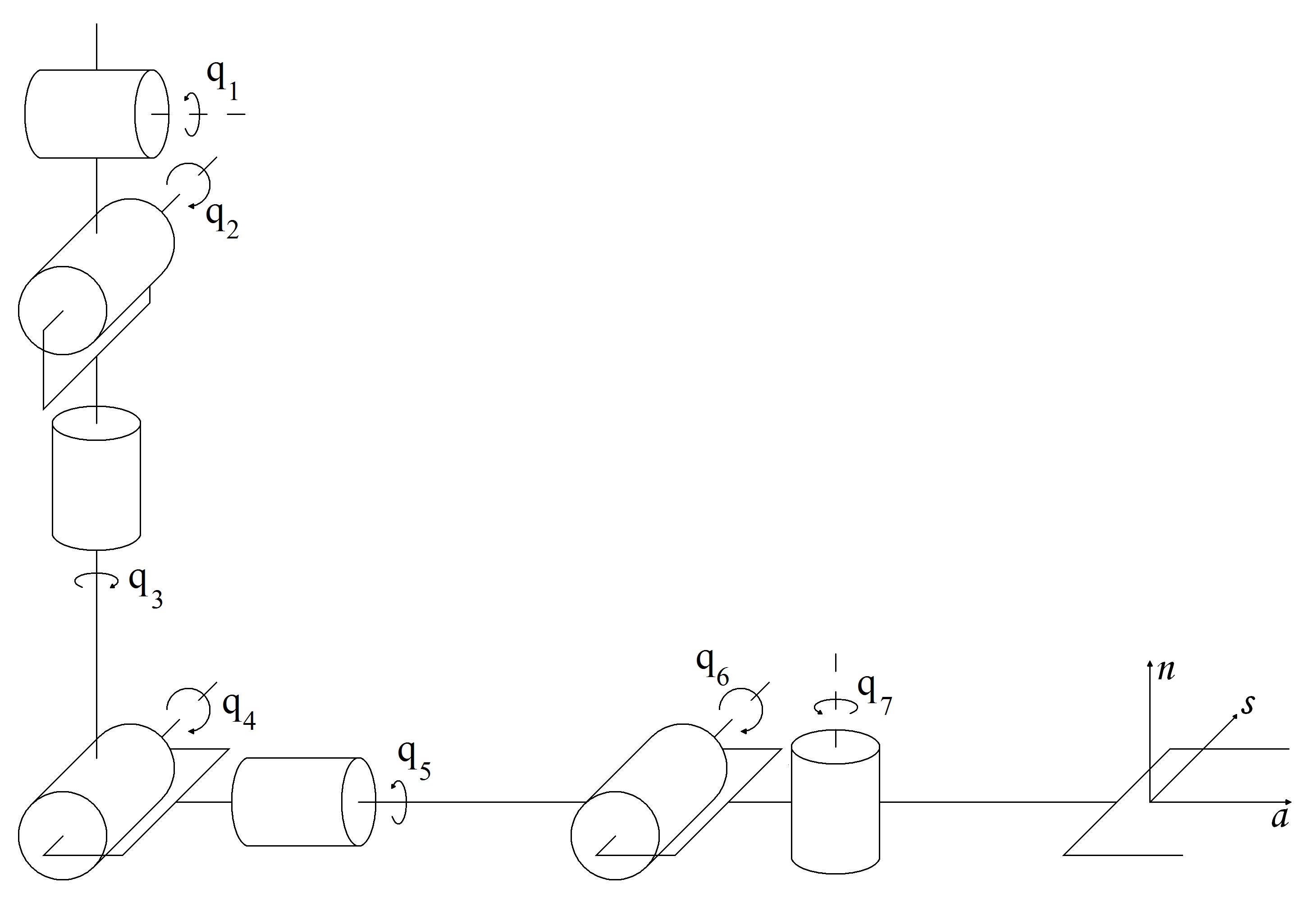}
  \caption{Left: Graphical representation of the PERA \cite{Rijs2010philips}. Right: Denavit-Hartenberg representation of the PERA \cite{Munoz2015energy}.}
  \label{fig-PERA}
\end{figure}

\subsection{End-effectors formation in 3D}
This subsection provides simulation results using $N=4$ Philips Experimental Robot Arms (PERA) in 3D space.
As depicted in Fig.~\ref{fig-PERA}, the PERA has seven DOF, and its dynamic model and Denavit-Hartenberg representation can be found in \cite[Appendix A]{Munoz2015energy}.
The desired shape is a tetrahedron with slide length of 0.4 m. The incidence matrix is
\begin{equation*}
B = \begin{bmatrix} 1 & 1 & -1 & 1 & 0 & 0 \\ -1 & 0 & 0 & 0 & 0 & 1 \\ 0 & -1 & 1 & 0 & 0 & -1 \\ 0 & 0 & 0 & -1 & -1 & -1 \end{bmatrix}.
\end{equation*}

The bases of the 4 manipulators are located at $(0,0)$, $(0.5,0)$, $(0.5,0.5)$ and $(0,0.5)$, respectively.
Using the distributed formation control as presented in Theorem~\ref{thm-III}, and following the parameter estimation method in the previous example, we set the controller parameters as follows: $\alpha=0.01$, $K_{P}=120$, $K_{I}=1$ and $K_{D}=20$.

\begin{figure}
  \centering
  \includegraphics[width=0.5\textwidth]{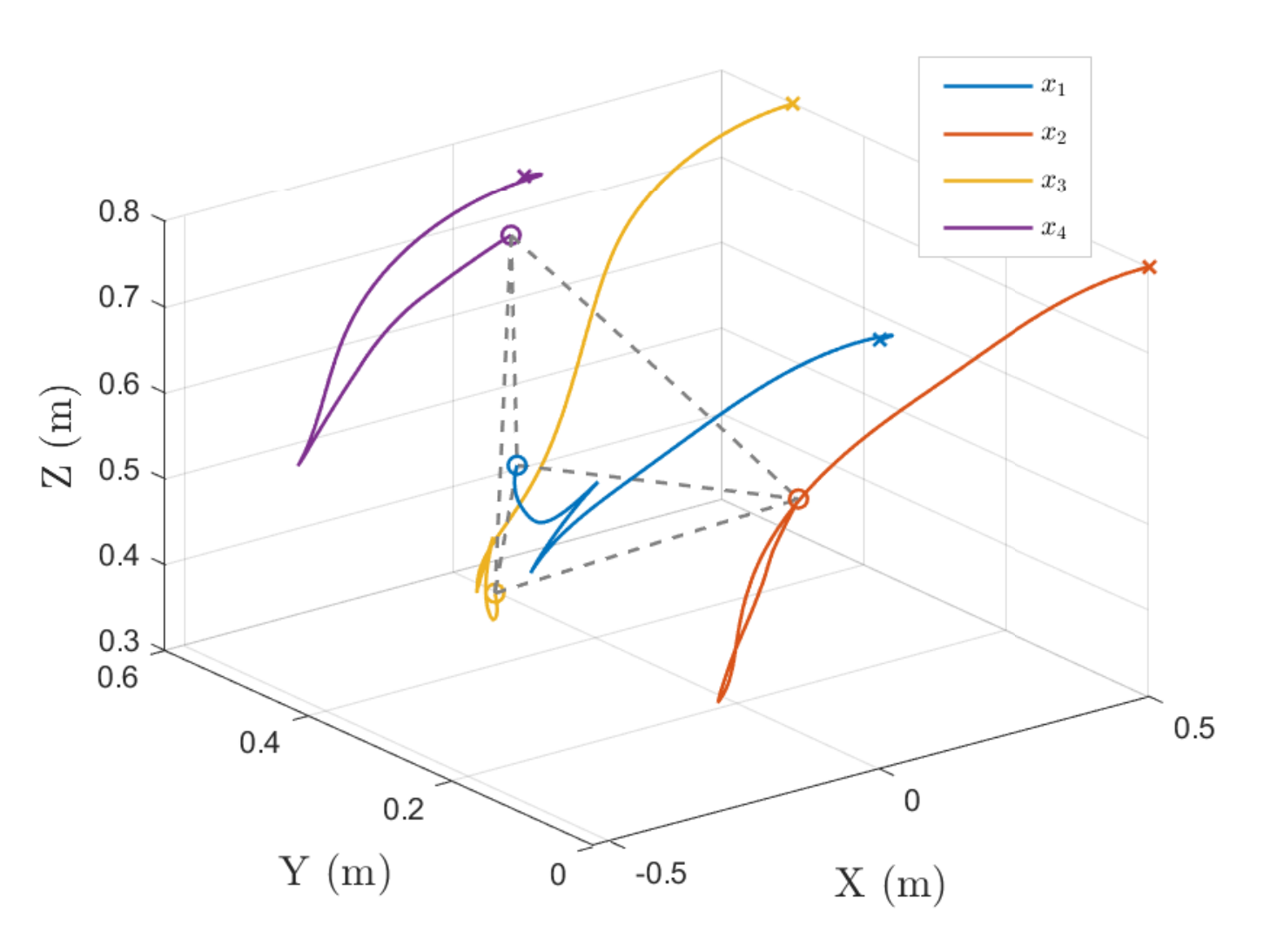}
  \caption{Trajectories of the manipulator end-effectors from the initial positions ($\times$) to the final positions ($\circ$) in 3D space.}\label{fig-x-3D}
\end{figure}

\begin{figure}
  \centering
  \includegraphics[width=0.5\textwidth]{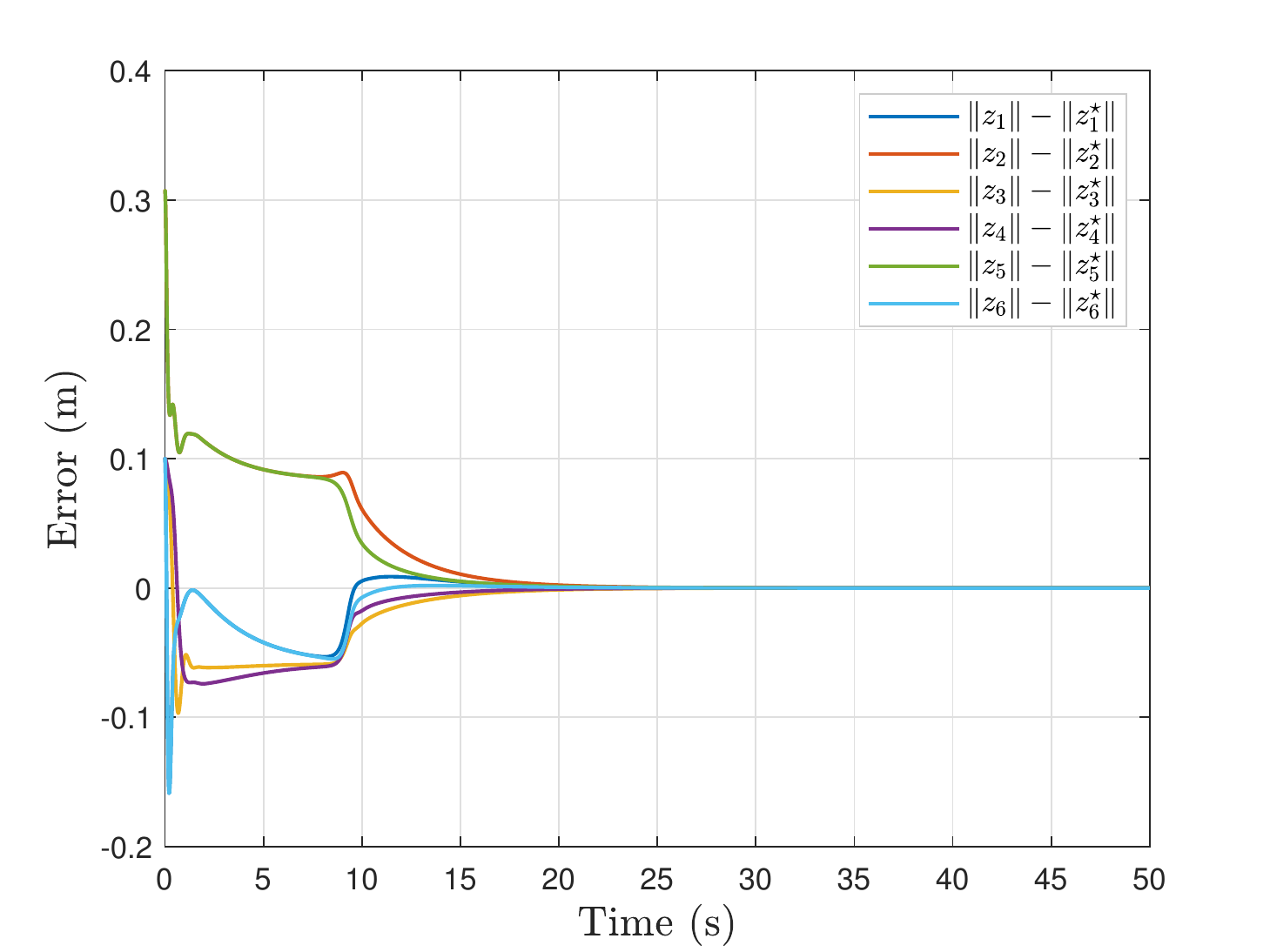}
  \caption{Performance of the inner distance error in 3D space.}\label{fig-e-3D}
\end{figure}

\begin{figure}
  \centering
  \includegraphics[width=0.5\textwidth]{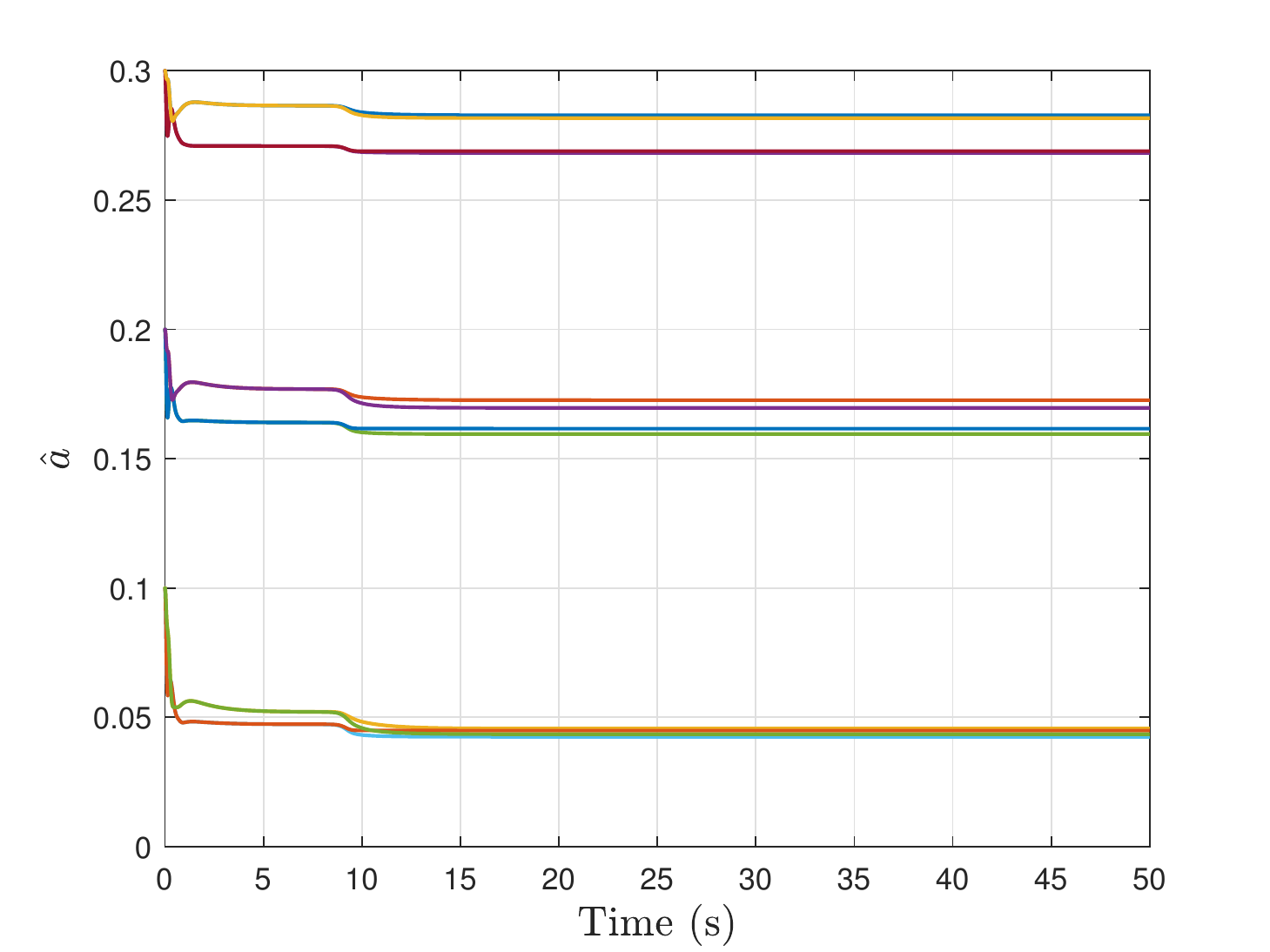}
  \caption{Kinematic parameter estimates in 3D space.}\label{fig-est-3D}
\end{figure}

\begin{figure}
  \centering
  \includegraphics[width=0.5\textwidth]{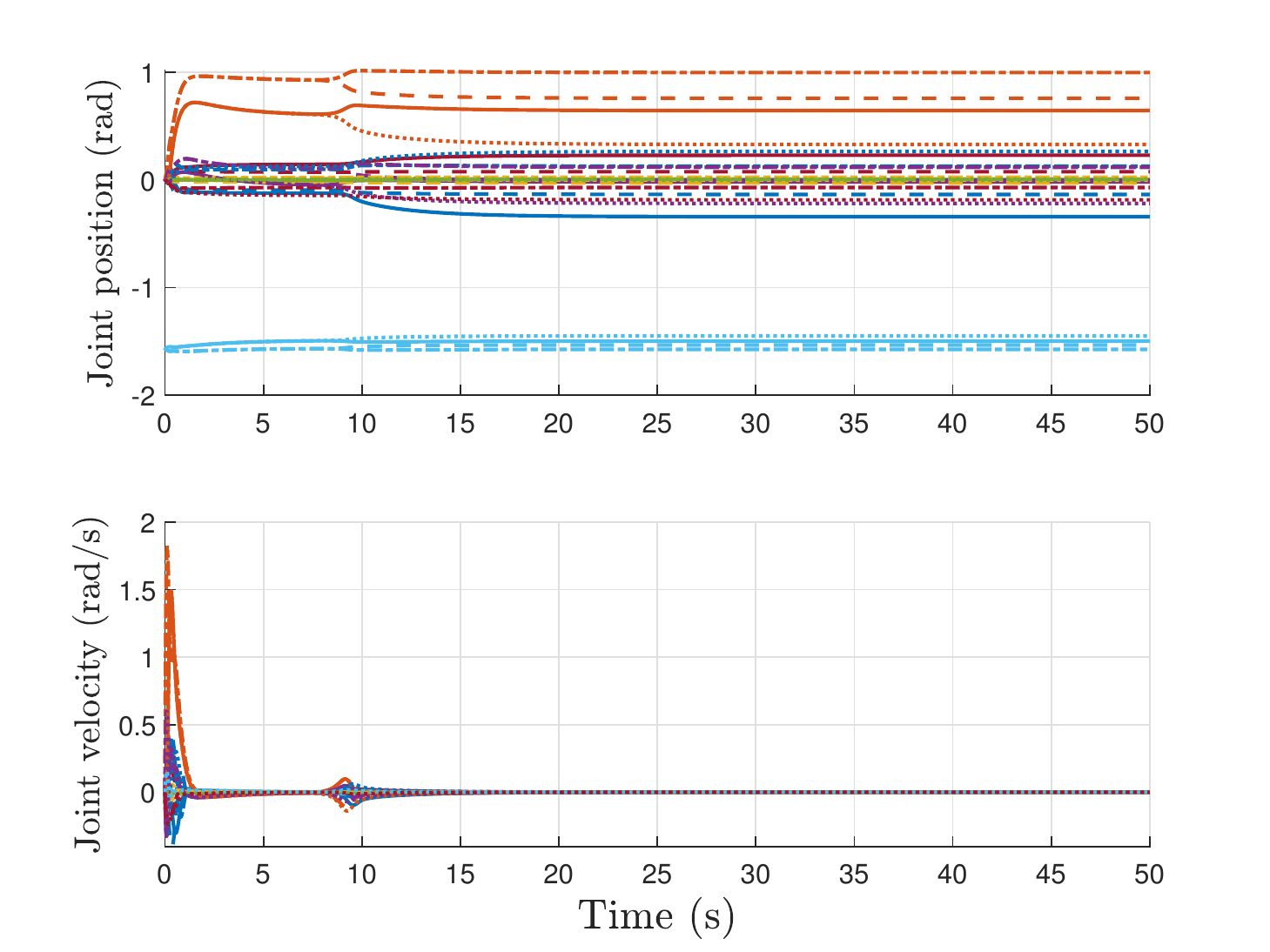}
  \caption{Performance of joint trajectories and velocities in 3D space.}\label{fig-joint-3D}
\end{figure}

Based on this simulation setup, we run the simulation for $50$ seconds until the formation converges and the simulation results are shown in Figures \ref{fig-x-3D} to \ref{fig-joint-3D}. The trajectories and formation pattern of the manipulators' end-effector as presented
in Fig.~\ref{fig-x-3D}. Fig.~\ref{fig-e-3D} shows that the inner distance errors converge to zero as expected.
Fig.~\ref{fig-est-3D} the evaluation of kinematic parameter estimates.
From Fig.~\ref{fig-joint-3D}, where the joint positions and velocities are plotted, we can conclude that the end-effectors remain stationary once they reach the intended shape, e.g., they do not exhibit undesirable group motion.

\section{Conclusion}\label{sec-con}
We have presented and analyzed gradient descent-based distributed formation controllers for end-effectors.
By introducing an extra integrator and an adaptive estimator for gravitational compensation and stabilization, respectively, we extend the proposed gradient-based design to the case where the manipulator kinematic and dynamic parameters are not exactly known.
The efficacy of the proposed methods is shown in simulation.

\appendices
\section*{Appendix}

\subsection{Upper bound of $\phi_{1}(t,e,\xi,w)$}\label{app-phi-1}
By using \eqref{J-dt1} and $\widehat{e} = 2\overline{B}D_{z}e$, we can rewrite $\phi_{1}(t,e,\xi,w)$ defined in \eqref{phi-1} as follows
\begin{align*}
&\phi_{1}(t,e,\xi,w) \\
&= 2e^{T}\underbrace{D_{(\overline{B}^{T}J(q(t),a)\xi)}^{T}\overline{B}^{T}J(q(t),a)H(q(t),w)}_{f_{11}(q(t),\xi,w)}\xi \\
&\quad + 4\xi^{T}\underbrace{J^{T}(q(t),a)\overline{B}D_{z} D_{z}^{T}\overline{B}^{T}J(q(t),a)H(q(t),w)}_{f_{12}(q(t),e,w)}\xi \\
&\quad + 2e^{T}\underbrace{D_{z}^{T}\overline{B}^{T}\dot{J}(q(t),a)H(q(t),w)}_{f_{13}(q(t),e,w)}\xi \\
&\quad + 2e^{T}\underbrace{D_{z}^{T}\overline{B}^{T}J(q(t),a)C^{T}(q(t),\xi,w)}_{f_{14}(q(t),e,\xi,w)}\xi.
\end{align*}
Let $\beta_{11}$, $\beta_{12}$, $\beta_{13}$ and $\beta_{14}$ be the maximum induced 2-norm for functions $f_{11}$, $f_{12}$, $f_{13}$ and $f_{14}$ in $\{(q,e,\xi,w): q\in\R^{nN},\|e\|^{2} \leq r_{1}, \|\xi\|^{2} \leq r_{2}, w\in\mathbb{W}\}$, respectively.

By using Young's inequality to the cross terms, we can bound $\phi_{1}$ as follows
\begin{align*}
\|\phi_{1}(t,e,\xi,w)\| &\leq (\beta_{11} + \beta_{13} + \beta_{14}) \|e\|^{2} \\
&\quad + (\beta_{11} + 4\beta_{12} + \beta_{13} + \beta_{14}) \|\xi\|^{2}
\end{align*}
which confirms \eqref{phi1-Young} with $k_{11} = \beta_{11} + \beta_{13} + \beta_{14}$ and $k_{12} = \beta_{11} + 4\beta_{12} + \beta_{13} + \beta_{14}$.

\subsection{Upper bound of $\dot{V}_{\eta}|_{\eqref{im-e}} $}
The time derivative of $V_{\eta}$ defined in \eqref{V-eta} along the trajectory of $\eqref{im-e}$ satisfies
\begin{align}\label{V-eta-app}
\dot{V}_{\eta}|_{\eqref{im-e}} &= -\tilde{\eta}^{T}\tilde{\eta} + \tilde{\eta}^{T}\big[- H(q(t),w)\xi - K_{I}^{-1}C^{T}(q(t),\xi,w)\xi \nnum\\
&\quad + K_{I}^{-1}G(q(t),w) - K_{I}^{-1}G(q^{*},w)\big] \nnum\\
&=  -\tilde{\eta}^{T}\tilde{\eta} + \tilde{\eta}^{T} \underbrace{\left[K_{I}^{-1}G(q(t),w) - K_{I}^{-1}G(q^{*},w) \right]}_{f_{21}(q(t),q^{*},w)} \nnum\\
&\quad - \tilde{\eta}^{T} \underbrace{\left[H(q(t),w) + K_{I}^{-1}C^{T}(q(t),\xi,w)\right]}_{f_{22}(q(t),\xi,w)}\xi .
\end{align}
To show $\|f_{21}(q(t),q^*,w)\|$ can be bounded by a function of $\|e\|$, we first present the relationship between $q$ and $e$
\begin{align*}
e_{k} &= \|z_{k}\|^{2} - \|z_{k}^{*}\|^{2} \\
&= \|x_{i} - x_{j}\|^{2} - \|x_{i}^{*} - x_{j}^{*}\|^{2} \\
&= \underbrace{[(x_{i} - x_{j}) + (x_{i}^{*} - x_{j}^{*})]^{T}}_{\chi_{k}^{T}}[(x_{i} - x_{i}^{*}) - (x_{j}- x_{j}^{*})]
\end{align*}
where the last equality is obtained by using the equality $\|a\|^{2} - \|b\|^{2} = a^{T}a - b^{T}b = (a + b)^{T}(a - b)$, $\forall a,b\in\R^{m}$. Then, we can write down $e$ in the compact form
\begin{align*}
e &= D_{\chi}^{T}\overline{B}^{T}(x - x^{*}) = D_{\chi}^{T}\overline{B}^{T}[h(q(t),w) - h(q^{*},w)]
\end{align*}
where $D_{\chi} = \textnormal{block diag}(\chi_{1},\dots,\chi_{|\E|})$, $x = \col(x_{1},\dots,x_{N})$ and $x^{*} = \col(x_{1}^{*},\dots,x_{N}^{*})$.
It can be verified that $\chi_{i}^{T}\chi_{j}, (i,j)\in\E$ can be written as a function of $e$. Since the graph $\G$ is infinitesimally and minimally rigid, $D_{\chi}^{T}\overline{B^{T}B}D_{\chi}$ is positive definite at $e = \mathbf{0}$, and $D_{\chi}^{T}\overline{B}^{T}$ is full rank in the set $\{e:\|e\|^{2} \leq r\}$ for some $r>0$. Thus, there exist $\kappa_{1} > 0$ such that $\|h(q(t),w) - h(q^{*},w)\| \leq \kappa_{1}\|e\|$
holds for all admissible $q(t)$, all $e\in\mathcal{Q}$ and all $w\in\mathbb{W}$.
Since $h(q(t),w)$ and $G(q(t),w)$ are smooth functions depending on $q(t)$ as arguments of bounded trigonometric functions for revolute joint manipulators, there exist $\kappa_{2}>0$ such that $\|G(q(t),w) - G(q^{*},w)\| \leq \kappa_{2}\|h(q(t),w) - h(q^{*},w)\| $
for all $q(t)\in\R^{nN}$ and $w\in\mathbb{W}$.
Therefore, once $K_{I}$ is chosen, we have
\begin{equation}\label{f12}
\|f_{12}(q(t),q^{*},w)\| \leq \beta_{21}\|e\|
\end{equation}
with $\beta_{21} = \kappa_{1}\kappa_{2}K_{I}^{-1}$, for all admissible $q(t)$, all $e\in\mathcal{Q}$ and all $w\in\mathbb{W}$.

Let $\beta_{22}$ be the maximum induced 2-norm for functions $f_{22}$ in $\{(q,\xi,w): q\in\R^{nN}, \|\xi\|^{2} \leq r_{2}, w\in\mathbb{W}\}$.

Then, by Young's inequality to the cross term in \eqref{V-eta-app}, we obatin
\begin{align*}
\dot{V}_{\eta}|_{\eqref{im-e}} &\leq - \|\tilde{\eta}\|^{2} + \frac{1}{4}\|\tilde{\eta}\|^{2} + \|f_{21}(q(t),q^{*},w)\|^{2} \\
&\quad + \frac{1}{4}\|\tilde{\eta}\|^{2} + \|f_{22}(q(t),\xi,w)\|^{2}\|\xi\|^{2} \\
&\leq - \frac{1}{2}\|\tilde{\eta}\|^{2}  + \beta_{21}^{2}\|e\|^{2} + \beta_{22}^{2}\|\xi\|^{2} 
\end{align*}
It validates \eqref{im-e} with $k_{21} = \beta_{21}^{2}$ and $k_{22} = \beta_{22}^{2}$.

\subsection{Upper bound of $\phi_{21}(t,\tilde{\eta},e,\xi,w)$ and $\phi_{22}(t,\tilde{\eta},e,\xi,w)$}
In view of $K_{I}\eta^{*} = G(q^{*},w)$, we can write 
\begin{align*}
\phi_{21}(t,\tilde{\eta},e,\xi,w) &= \xi^{T}K_{I}\tilde{\eta} + \xi^{T}K_{I}H(q(t),w)\xi \\
&\quad + \xi^{T}[G(q^{*},w) - G(q(t),w)]
\end{align*}
Then, in a similar manner as the above, we have
\begin{align*}
\|\phi_{21}(t,\tilde{\eta},e,\xi,w)\| &\leq K_{I} (\frac{1}{2}\|\xi\|^{2} + \frac{1}{2}\|\tilde{\eta}\|^{2}) + K_{I}c_{\max}\|\tilde{\eta}\|^{2} \\
&\quad +K_{I}(\frac{1}{2}\|\xi\|^{2} + \frac{1}{2}\|f_{12}(q(t),q^{*},w)\|^{2}).
\end{align*}
Substituting \eqref{f12} into the above yields
\begin{align*}
\|\phi_{21}(t,\tilde{\eta},e,\xi,w)\| &\leq K_{I}\|\xi\|^{2} + \frac{1}{2}K_{I}\beta_{12} \|e\|^{2} \\
&\quad + (\frac{1}{2}K_{I} + K_{I}c_{\max})\|\tilde{\eta}\|^{2}
\end{align*}
Hence, the inequality \eqref{phi22-Young} holds with $k_{31}=\frac{1}{2}K_{I} + K_{I}c_{\max}$, $k_{32} = \frac{1}{2}K_{I}\beta_{12}$ and $k_{33} = K_{I}$.

\bigskip

Next, by using \eqref{phi-1}, function $\phi_{22}(t,e,\xi,w)$ satisfies
\begin{align*}
\phi_{22}(t,e,\xi,w)
&= \phi_{1}(t,e,\xi,w) + 2e^{T}D_{z}^{T}\overline{B}^{T}[K_{I}\tilde{\eta} \\
&\quad + K_{I}H(q(t),w)\xi + G(q^{*},w) - G(q(t),w)] \\
&= \phi_{1}(t,e,\xi,w) + 2e^{T}D_{z}^{T}\overline{B}^{T}K_{I}\tilde{\eta} \\
&\quad + 2e^{T}D_{z}^{T}\overline{B}^{T}K_{I}H(q(t),w)\xi \\
&\quad + 2e^{T}D_{z}^{T}\overline{B}^{T}K_{I}f_{12}(q(t),q^{*},w)
\end{align*}
Let $\beta_{31}$ be the maximum induced 2-norm for $D_{z}^{T}\overline{B}^{T}$ in $\{e:\|e\|^{2}\leq r_{1}\}$.
Then, by using the Young's inequality and using \eqref{f12}, we have
\begin{align*}
\|\phi_{22}(t,e,\xi,w)\|
&\leq \|\phi_{1}(t,e,\xi,w)\| + \beta_{31}K_{I}(\|e\|^{2} + \|\tilde{\eta}\|^{2}) \\
&\quad + \beta_{31}K_{I}c_{\max} (\|e\|^{2} + \|\xi\|^{2}) \\
&\quad + \beta_{31}K_{I} \beta_{21} \|e\|^{2}.
\end{align*}
Finally, substituting \eqref{phi-1} into the above, we obtain \eqref{phi22-Young} with $k_{41} = \beta_{31}K_{I}$, $k_{42} = k_{11}+\beta_{31}K_{I}+\beta_{31}K_{I}c_{\max}+\beta_{31}K_{I} \beta_{21}$ and $k_{43}= k_{12}+\beta_{31}K_{I}c_{\max}$.


\ifCLASSOPTIONcaptionsoff
  \newpage
\fi

\bibliographystyle{IEEEtran}
\bibliography{robotBIB}




\begin{IEEEbiography}[{\includegraphics[width=1in,height=1.25in,clip,keepaspectratio]{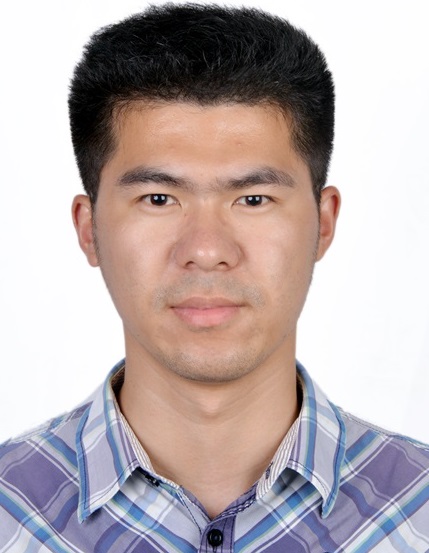}}]{Haiwen Wu}
received the B.Sc. degree in automatic control from Nanjing University of Science and Technology, China, in 2014.
He is currently with the Faculty of Science and Engineering, University of Groningen, The Netherlands, working toward his Ph.D. degree.
His research interest includes nonlinear control, output regulation, and robotic systems.
\end{IEEEbiography}

\begin{IEEEbiography}[{\includegraphics[width=1in,height=1.25in,clip,keepaspectratio]{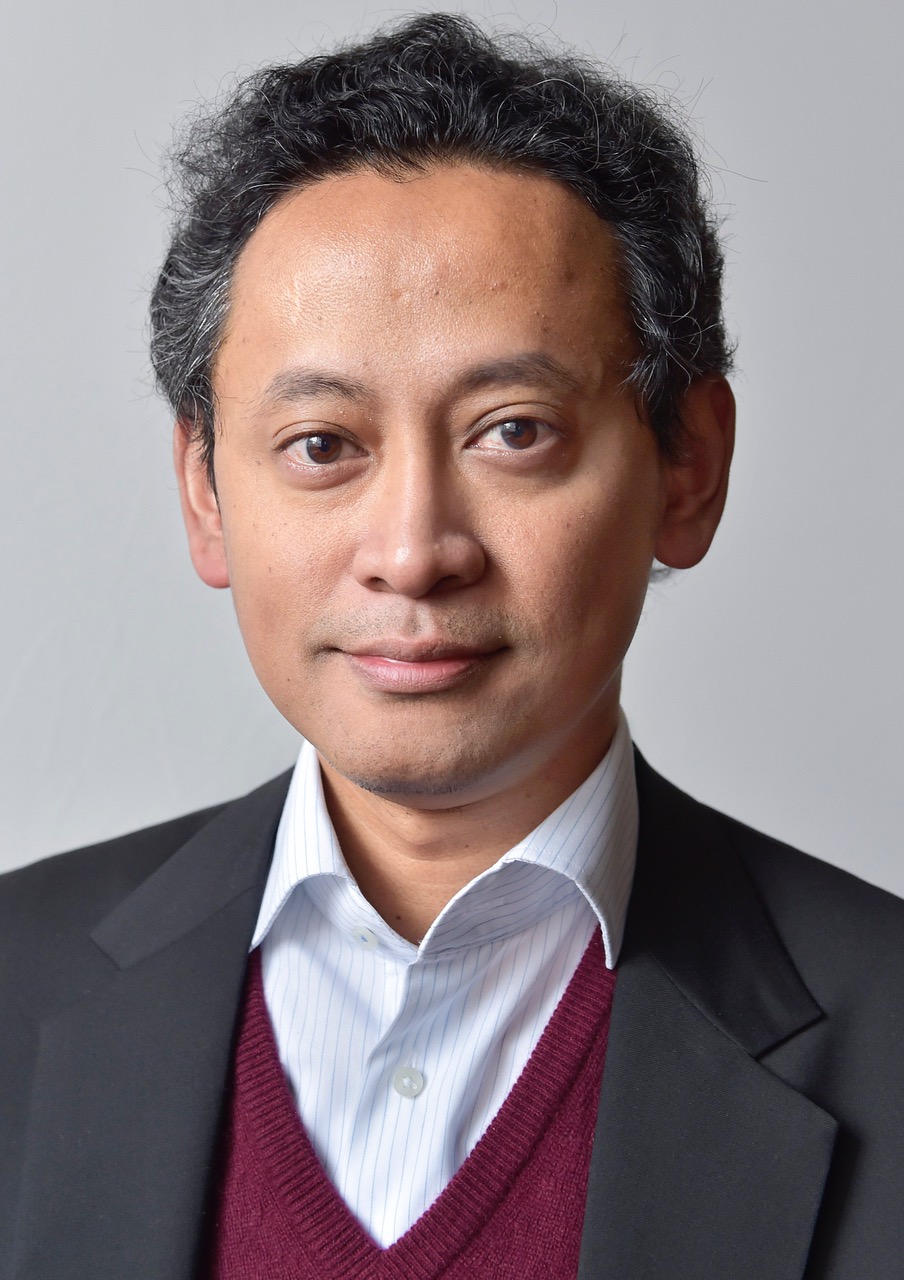}}]{Bayu Jayawardhana}
(SM’13) received the B.Sc. degree in electrical and electronics engineering from the Institut Teknologi Bandung, Bandung, Indonesia, in 2000, the M.Eng. degree in electrical and electronics engineering from the Nanyang Technological University, Singapore, in 2003, and the Ph.D. degree in electrical and electronics engineering from Imperial College London, London, U.K., in 2006. He is currently a professor of mechatronics and control of nonlinear systems in the Faculty of Science and Engineering, University of Groningen, The Netherlands. He was with Dept. Mathematical Sciences, Bath University, Bath, U.K., and with Manchester Interdisciplinary Biocentre, University of Manchester, Manchester, U.K. His research interests include the analysis of nonlinear systems, systems with hysteresis, mechatronics, systems and synthetic biology. Prof. Jayawardhana is a Subject Editor of the International Journal of Robust and Nonlinear Control, an Associate Editor of the European Journal of Control and a member of the Conference Editorial Board of the IEEE Control Systems Society.
\end{IEEEbiography}

\begin{IEEEbiography}[{\includegraphics[width=1in,height=1.25in,clip,keepaspectratio]{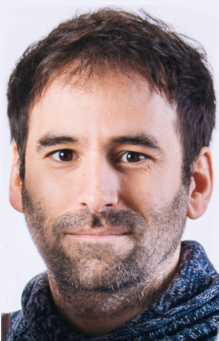}}]{Hector Garcia de Marina}
(M’17) received the M.Sc. degree in electronics engineering from the Complutense University of Madrid, Madrid, Spain, in 2008, the M.Sc. degree in control engineering from the University of Alcala, Alcala de Henares, Spain, in 2011, and the Ph.D. degree in systems and control from the University of Groningen, Groningen, the Netherlands, in 2016. He held a postdoctoral position in the \'{E}cole Nationale de l’Aviation Civile in Toulouse from 2016 to 2018. From 2018 to 2020, he was an assistant professor at the Unmanned Aerial Systems Center, University of Southern Denmark. Since 2020, he currently holds a research fellow in the Complutense University of Madrid. His research interests include the guidance navigation and control for autonomous robots, and multi-agent systems.
\end{IEEEbiography}

\begin{IEEEbiography}[{\includegraphics[width=1in,height=1.25in,clip,keepaspectratio]{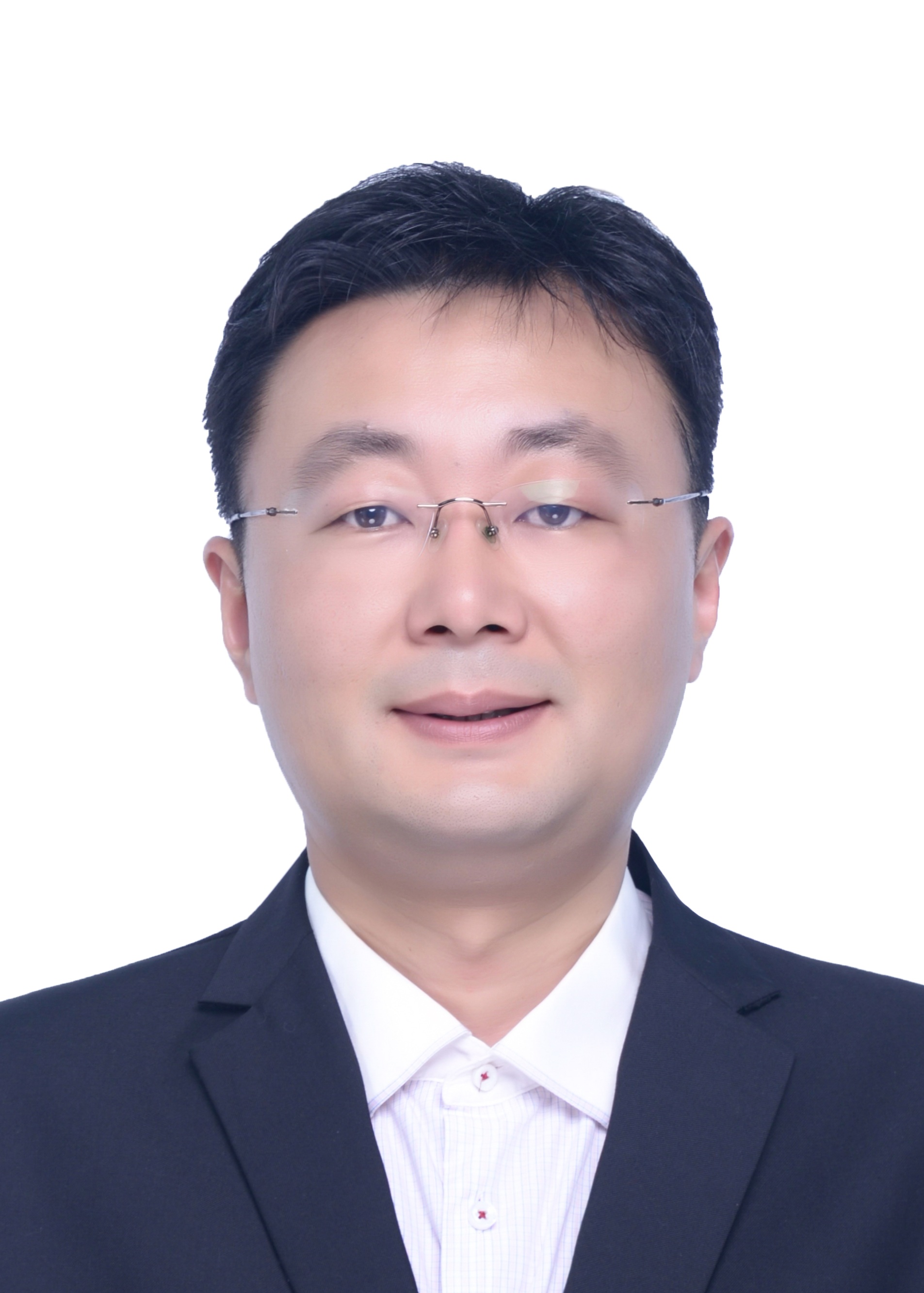}}]{Dabo Xu}
received the B.Sc. degree in mathematics and applied mathematics from Qufu Normal University, China, in 2003, the M.Sc. degree in operations research and cybernetics from Northeastern University, China, in 2006, and the Ph.D. degree in automation and computer-aided engineering from The Chinese University of Hong Kong, Hong Kong, China, in 2010. He is currently a professor at School of Automation, Nanjing University of Science and Technology, China. He was a postdoctoral fellow at The Chinese University of Hong Kong and then a research associate at The University of New South Wales at Canberra, Australia. His current research focus is on nonlinear control and distributed control with their applications to modeling and control of robotic manipulators and unmanned aerial vehicles. He is a Subject Editor of the International Journal of Robust and Nonlinear Control, a member of the editorial board of Journal of Systems Science and Complexity, and an Associate Editor of Control Theory and Technology.
\end{IEEEbiography}

\vfill

\end{document}